\documentclass[runningheads]{llncs}

\usepackage[T1]{fontenc}

\usepackage{amsmath,amssymb,amsfonts}
\usepackage{algorithmic}
\usepackage{graphicx}
\usepackage[ruled, vlined]{algorithm2e}
\usepackage{textcomp}
\usepackage{multirow}
\usepackage{enumitem}
\usepackage{hyperref}
\usepackage{caption}
\usepackage{pdflscape}
\usepackage{outlines}
\usepackage{gensymb}
\usepackage{soul}
\usepackage{booktabs}  
\usepackage{subcaption}
\usepackage{csquotes}
\usepackage[dvipsnames]{xcolor}
\usepackage{wrapfig}
\usepackage{siunitx}
\usepackage{adjustbox}
\usepackage[misc,geometry]{ifsym}
\usepackage{dblfloatfix}
\usepackage{float}
\usepackage{subcaption}

\usepackage[maxbibnames=2,doi=false,isbn=false,url=false, backend=biber, style=nature]{biblatex}
\usepackage{balance}

\begin{document}

\newcommand{\paper}{SplitGuardian}

\title{SplitOut: Out-of-the-Box Training-Hijacking Detection in Split Learning via Outlier Detection}
\titlerunning{SplitOut}

\author{Ege Erdoğan\textsuperscript{(\Letter)}\inst{1}\and
Unat Tekşen\inst{2}\and
M. Salih Çeliktenyıldız\inst{3}\and
Alptekin Küpçü\inst{4}\and
A. Ercüment Çiçek\inst{3}
}
\authorrunning{E. Erdogan et al.}

\institute{
Technical University of Munich, Munich, Germany
\email{ege.erdogan@tum.de}
\and
Kadir Has University, Istanbul, Turkey
\email{20181701048@stu.khas.edu.tr}\\
\and
Bilkent University, Ankara, Turkey
\email{m.celiktenyildiz@ug.bilkent.edu.tr, cicek@cs.bilkent.edu.tr}\\
\and
Koç University, Istanbul, Turkey
\email{akupcu@ku.edu.tr}\\
}

\maketitle 

\begin{abstract}
Split learning enables efficient and privacy-aware training of a deep neural network by splitting a neural network so that the clients (data holders) compute the first layers and only share the intermediate output with the central compute-heavy server. This paradigm introduces a new attack medium in which the server has full control over what the client models learn, which has already been exploited to infer the private data of clients and to implement backdoors in the client models. Although previous work has shown that clients can successfully detect such training-hijacking attacks, the proposed methods rely on heuristics, require tuning of many hyperparameters, and do not fully utilize the clients' capabilities. In this work, we show that given modest assumptions regarding the clients' compute capabilities, an out-of-the-box outlier detection method can be used to detect existing training-hijacking attacks with almost-zero false positive rates. We conclude through experiments on different tasks that the simplicity of our approach we name \textit{SplitOut} makes it a more viable and reliable alternative compared to the earlier detection methods.

\end{abstract}


\section{Introduction}\label{sec:introduction}

Neural networks often perform better when they are trained on more data using more parameters, which in turn requires more computational resources. However, the required computing resources might be inaccessible in many application scenarios, and it is not always possible to share data freely in fields such as healthcare \cite{mercuri_hipaa-potamus_2004}. 

Distributed and/or outsourced deep learning frameworks such as \textit{split learning} \cite{vepakomma_split_2018, gupta_distributed_2018} aim to solve these problems by enabling a group of resource-constrained data holders (clients) to collaboratively train a neural network on their collective data without explicitly sharing private data, and also while offloading a (large) share of computational load to more resourceful servers. 

In split learning, a neural network is split into two parts, with the clients computing the first few layers on their private input and sharing the output with the central server, who then computes the rest of the layers. 
While SplitNN can indeed utilize the collective data set of the clients without raw data sharing, SplitNN and other distributed training frameworks have been shown to leak information about private data in different ways \cite{erdogan2021unsplit, pasquini_unleashing_2021, gao2023pcat, fredrikson2015model, zhu2019deep, fuFocusingPinocchioNose2023}.


\textbf{The Problem}. What a split learning client learns is fully determined by the server, since the clients' parameter updates are propagated back from the gradients the server sends. A malicious server can thus launch \textit{training-hijacking} attacks that have been shown to be effective in inferring clients' private data \cite{pasquini_unleashing_2021, fuFocusingPinocchioNose2023} and implementing backdoors in the client models \cite{yuHowBackdoorSplit2023}. Nevertheless, since training is an iterative process, clients can potentially detect training-hijacking attacks and stop training before the attack converges. Two such detection methods have been proposed in earlier work: an active method that relies on the clients tampering with the data to observe the resulting changes in the system \cite{splitguard}, and a passive method that depends only on observing the system \cite{fuFocusingPinocchioNose2023}, motivated by the fact that active methods can be detected and circumvented by the attacker. Both of these methods primarily rely on heuristic approaches that do not fully utilize the clients' capabilities, which motivates the search for less fragile and simpler-to-use approaches.  

\begin{figure*}[t!]
    \centering
    \begin{subfigure}{0.48\textwidth}
        \centering
        \includegraphics[width=\textwidth]{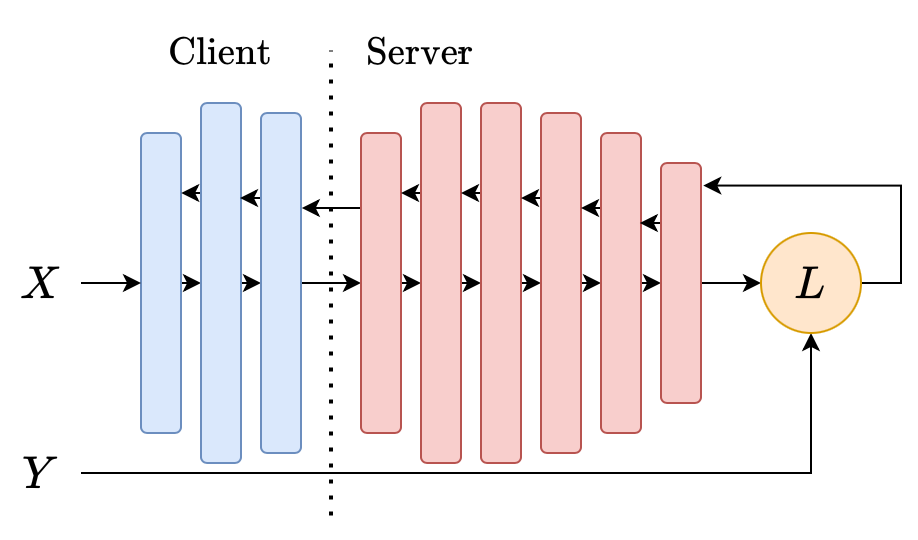}
        \caption{With label-sharing.}
        \label{fig:splitnn_label_sharing}
    \end{subfigure}  
    \begin{subfigure}{0.48\textwidth}
        \centering
        \includegraphics[width=\textwidth]{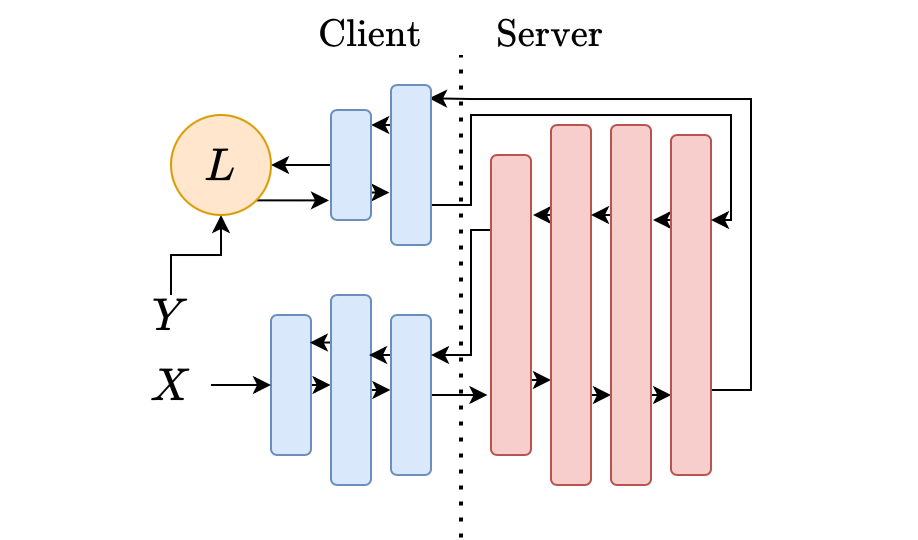}
        \caption{Without label-sharing.}
        \label{fig:splitnn_private_labels}
    \end{subfigure}
    \caption{\textbf{Potential SplitNN setups}. Arrows denote the forward and backward passes, starting with the input data $X$, and propagating backwards after the loss computation using the labels $Y$. In Figure \ref{fig:splitnn_label_sharing}, clients send the labels to the server along with the intermediate outputs. In Figure \ref{fig:splitnn_private_labels}, the model terminates on the client side, and thus the clients do not share their labels.}
    \label{fig:splitnn_setups}
\end{figure*}


\textbf{Our Solution}. We show that given modest assumptions on the clients' compute capabilities, an out-of-the-box outlier detection algorithm can successfully distinguish the gradients sent by a malicious training-hijacking server from those sent by an honest one with almost perfect accuracy, resulting in a simple-yet-effective passive detection method. We empirically demonstrate the effectiveness of the method against the existing training-hijacking attacks \cite{pasquini_unleashing_2021, yuHowBackdoorSplit2023, fuFocusingPinocchioNose2023} and an adaptive attacker that can bypass the active method of \cite{splitguard}. As a further contribution beyond the method proposed, through the data we have collected during the experiments, we discover that the differences between the honest training and training-hijacking optimization processes result in training-hijacking gradients having significantly distinguishable neighborhood characteristics compared to honest gradients, which explains the effectiveness of the neighborhood-based local outlier factor outlier detection algorithm \cite{breunig2000lof} that can be used out-of-the-box from popular frameworks \cite{scikit-learn}. Due to the proactive design of our solution, we think that it can potentially detect some not-yet-known attacks, rather than taking a reactive approach.

Our work thus results in a unifying framework of training-hijacking attacks in split learning, and leads to new insights on their behavior. We make our code anonymously available at \url{https://github.com/ege-erdogan/splitout}.


\section{Background and Motivation}

\subsection{Neural Networks and Data Privacy} \label{sec_nn_priv}

Modern machine learning (ML) methods employ the principle of \textit{empirical risk minimization} \cite{vapnik1991principles}, where the aim is to find a function $f$ with parameters $\theta$ that minimize some loss over a dataset. Many modern systems utilize variants of the iterative \textit{stochastic gradient descent} (SGD) method, where parameters are updated by sampling a batch of data and taking a step in the direction of steepest descent on the loss landscape.

Neural networks \cite{rosenblatt1958perceptron} define a certain kind of function space. In its simplest form, each neuron in a neural network layer computes a weighted sum of the previous layer neurons and applies a non-linear \textit{activation}, such as the sigmoid function. Single-layer neural networks defined in this way are universal approximators \cite{hornik1991approximation} provided they are wide enough; however, stacking narrower layers to build \textit{deep neural networks} (DNNs) leads to more efficient training for the same number of parameters. Practical training of DNNs was made possible by the \textit{backpropagation algorithm} \cite{rumelhart1986learning}, in which a DNN is treated as a composition of functions and the gradient $\nabla_\theta \mathcal{L}$ is decomposed via the chain rule, starting with the final layer and flowing backwards through the network.

Although such defined \textit{fully-connected} DNNs are highly expressive, this expressivity results in huge search spaces that are difficult to optimize. Instead, the function space is restricted through \textit{inductive biases} that make assumptions about the structure of the data, such as using convolutional neural networks \cite{lecun1995convolutional} that are invariant to translations of the features (e.g. edges in images), graph neural networks \cite{kipf2016semi, velivckovic2023everything} that are invariant to the order in which the nodes of the graph are processed, or attention-based transformers \cite{vaswani2017attention, devlin2018bert} that are designed to process sequential data such as text. 


The effectiveness of these specialized model architectures, along with access to ever more data and compute resources has gradually led to ground-breaking performance in tasks such as language modeling \cite{brown_language_2020} and image generation \cite{rombach2022high}. This has in turn caused the research community to consider ML systems in a more holistic way, focusing e.g. on their safety \cite{hendrycks2021unsolved} and energy use \cite{patterson2021carbon}, but most importantly for our purposes, \textit{data privacy}. 

Data privacy concerns and DNNs relate in two ways: 1) DNNs often perform better given more data, yet privacy regulations in fields such as healthcare \cite{mercuri_hipaa-potamus_2004} make it difficult to utilize existing data, and 2) neural networks, being the black-boxes they are, can leak training or input data \cite{carlini2023extracting, fredrikson2015model}. Data privacy has therefore been featured as a key concern in very recent regulatory work on artificial intelligence (AI) such as the Blueprint for an AI Bill of Rights published by the US Government \cite{White_House_2023} and the EU AI Act \cite{veale2021demystifying}. Attempts to enable privacy-aware training have resulted in frameworks such as \textit{split learning} which we focus on next.

\subsection{Split Learning} \label{sec_splitnn}

Split learning (SplitNN) \cite{gupta_distributed_2018, vepakomma_no_2018, vepakomma_split_2018} aims to address two challenges: 1) data-sharing in sensitive fields such as healthcare, and 2) training large models in environments with limited compute resources. The main idea is to split a DNN between clients (data holders) and a server. Each client locally computes the first few layers on their private data and sends the output to the server, who then computes the rest of the layers. This way, a group of clients can train a DNN utilizing, but not sharing, their collective data. This framework stands out as being more communication-efficient compared to other collaborative ML systems such as federated learning, especially with a higher number of clients and larger model sizes\cite{singh2019detailed}.

In our work, we model the clients as the data holders, and thus SplitNN has two possible settings depending on whether the clients share their labels with the server or not. If they do (Figure \ref{fig:splitnn_label_sharing}), the server computes the loss value, initiates backpropagation, and sends the gradients of its first layer back to the client, who then completes the backward pass. If, on the other hand, the clients do not share their labels (Figure \ref{fig:splitnn_private_labels}), the client computes the loss and there are two communication steps instead of one. Finally, clients take turns training with the server using their local data, and each client can obtain the most recent parameters from the last-trained client or through a central weight-sharing server. Interestingly due to this sequential training of clients, \textit{having many clients is logically equivalent to having a single client with all the data as the entire model is updated one sample (or mini-batch) at a time.} 

While a large part of computational work is transferred to the server, this involves a privacy/cost trade-off for the clients. The outputs of earlier layers are closer to the original inputs, and thus ``leak" more information. Choosing a split depth is therefore crucial for SplitNN to actually provide data privacy, as has been demonstrated in such attacks \cite{erdogan2021unsplit, pasquini_unleashing_2021, gao2023pcat}.

\subsection{Training-Hijacking in Split Learning} \label{sec_hijacking}

A split learning server has full control over the gradients sent to the initial client model and can thus direct it towards an adversarial objective of its choice. We call this attack medium \textit{training-hijacking}. It has already been exploited to infer the private data of the clients \cite{pasquini_unleashing_2021, fuFocusingPinocchioNose2023} and to implement backdoors in client models \cite{yuHowBackdoorSplit2023}.

Although the attack medium is not limited to this, all the attacks proposed so far perform \textit{feature-space alignment} between the client model and an adversarial \textit{shadow} model as an intermediate step. Then, once the client model's output space overlaps sufficiently with that of the shadow model, the downstream attack task can be performed. We first describe this alignment step and then explain how different attacks build on it. 

\subsubsection{Feature-Space Alignment}

First, we assume that the attacker has access to some dataset $X_{\text{pub}}$, which ideally follows the same distribution as the private client dataset $X_{\text{priv}}$, and the attacker's goal is to cause the outputs of the client model $f_c$ and the shadow model $f_s$ (potentially pre-trained on some task) to overlap. To this end, the attacker discards the original split learning training process to instead train a discriminator $D$ along with $f_s$ and $f_c$. Given an output from $f_s$ or $f_c$, the discriminator basically tries to distinguish which model it came from. Its objective is thus to minimize
\begin{equation} \label{eq:server_align}
    L_D = \log\ (1 - D \circ f_s (X_{\text{pub}})) \cdot (D\circ f_c(X_{\text{priv}}))
\end{equation}
while the client model $f_c$ is trained to maximize the discriminator's error; i.e. output values as similar to $f_s$ as possible, minimizing the loss function
\begin{equation} \label{eg:client_align}
    L_c = L(1 - D \circ f_c(X_{\text{priv}})).
\end{equation}
All training-hijacking attacks proposed as of this work \cite{pasquini_unleashing_2021, yuHowBackdoorSplit2023, fuFocusingPinocchioNose2023} share this alignment step, but they differ in how they train $f_s$ and the auxiliary models they implement to facilitate the attack. 

\subsubsection{Inferring Private Client Data}

To infer the private client data $X_{\text{priv}}$ using the feature-space alignment approach as in \cite{pasquini_unleashing_2021, fuFocusingPinocchioNose2023}, the shadow model $f_s$ is treated as the encoder half of an auto-encoder and the auxiliary model $f_a$ acts as the corresponding decoder; i.e. for some input $x$, we have $f_a \circ f_s (x) \approx x$. Crucially, the auto-encoder is trained on $X_{\text{pub}}$ before the feature-space alignment step in preparation for the attack. Then, provided that $X_{\text{pub}}$ and $X_{\text{priv}}$ are sufficiently similar, the combination $f_a \circ f_c$ also behaves as an auto-encoder on $X_{\text{priv}}$ and given a client output $z = f_c(x)$ for $x\in X_{\text{priv}}$, we have $f_a(z) \approx x$. This leads to an inversion attack. 

\subsubsection{Implementing Backdoors in the Client Model}

A model with a \textit{backdoor} outputs a different value for the same input when the input is equipped with a backdoor \textit{trigger}, e.g. an image classifier outputs ``dog" when the image is that of a cat but with a small white square added by the adversary \cite{wang2019neural}. For the attack to work, the model should be able to encode the inputs in a sufficiently different way when they are equipped with the trigger. To launch this attack within the training-hijacking paradigm, \cite{yuHowBackdoorSplit2023} assumes that the attacker again has access to some dataset $X_{\text{pub}}$ similar to $X_{\text{priv}}$ but with some of the samples containing the backdoor trigger. The shadow model $f_s$ is then combined with two models to form two classifiers: $f_m\circ f_s$ learns the original classification task, while $f_a \circ f_s$ learns the classify backdoor samples (i.e. binary classification), with the losses
\begin{align}
    L_m &= \mathcal{L}\left(f_m \circ f_s (X_{\text{pub}}, Y_{\text{pub}})\right) \\ 
    L_a &= \mathcal{L}\left(f_a \circ f_s (X_{\text{pub}}, B_{\text{pub}})\right) 
\end{align}
where $Y_{\text{pub}} \in \{1 ,..., C \}^N$ are the labels for the $C$ classes of the original classification task and $B_{\text{pub}} \in \{0,1\}^N$ denote whether an input contains the backdoor trigger or not. Intuitively, $f_s$ learns both the classification task (through $L_m$) and to encode the backdoor samples (through $L_a$). Then finally through the alignment process, $f_c$ behaves similarly to $f_s$ and also learns to encode the backdoor samples.

\subsection{Detecting Training-Hijacking Attacks} \label{sec_detecting}

Feature-space alignment is an iterative optimization process, and thus the attacks can be effectively prevented if the clients can detect it before the optimization converges. In the first attempt to do so, \cite{splitguard} observes that the objective in Equation \ref{eg:client_align} is independent of the label values unlike the classification objective. They propose a method called SplitGuard in which the clients randomly assign some label values and measure how the gradients received from the server differ between the randomized and non-randomized labels. 
If the server is malicious and is doing feature-space alignment instead of the original task, there should be no significant difference since the loss in Equation \ref{eg:client_align} does not include the label values. If, on the other hand, the server is honest, the gradients of the loss from the randomized labels should differ significantly from that of the non-randomized labels because success in the original task implies failure in the randomized task and vice versa. The gradients of the randomized labels will direct the model away from the points in the parameter-space that correspond to high performance in the non-randomized task.

This insight is used to compute the \textit{SplitGuard score} through the angle and the distances between the two sets of gradients. While the results in \cite{splitguard} indicate that SplitGuard can detect all instances of the feature-space hijacking attack of \cite{pasquini_unleashing_2021}, two weaknesses of the method are that it relies on choosing a number of hyperparameters during the computation of the score and the decision threshold, and that since it tampers with the training process, it can potentially be detected by the attacker and circumvented.


To overcome the second of these issues, \cite{fuFocusingPinocchioNose2023} proposes a passive detection mechanism relying on the observation that \textit{in a classification task, gradients of the samples with the same label should be more similar than those of the samples with different labels.} But since the discriminator loss in Equation \ref{eg:client_align} does not include the labels, this difference will be less visible if the server is performing feature-space alignment. 

Following a paradigm similar to SplitGuard \cite{splitguard}, \cite{fuFocusingPinocchioNose2023} aims to quantify this difference and apply a threshold-based decision rule. For each training step $i$, clients store the pair-wise cosine similarities of the gradients from samples with the same and different labels in $S_s(i)$ and $S_d(i)$, respectively. The quantification then consists of three components:
\begin{itemize}
    \item The \textit{set gap} $G_i$ computed as the mean difference between $S_s(i)$ and $S_d(i)$.
    \item The \textit{fitting error} (mean-square error) $E_n$ of quadratic polynomial approximations 
    \begin{align}
        P_s(i) &= a_s i^2 + b_s i + c_s \\
        P_d(i) &= a_d i^2 + b_d i + c_d
    \end{align}
    of $S_s(i)$ and $S_d(i)$, averaged across $n$ training iterations with $a,b,c$ the coefficients of the polynomial.
    \item The \textit{overlapping ratio} $V_i$ between $S_d(i)$ and $S_s(i)$ with the top and bottom $\gamma$ percentiles removed (denoted by $R_\gamma$), computed as
    \begin{equation}
        V_i = \frac{\vert R_\gamma(S_s(i)) \cap R_\gamma(S_d(i)) \vert}
                   {\vert R_\gamma(S_s(i)) \cup R_\gamma(S_d(i)) \vert}.
    \end{equation}
\end{itemize}
Finally, the final score for the $n$-th iteration is computed as
\begin{equation}
    \text{DS}_n = \text{Sigmoid}\big(\lambda_{\text{ds}}
        (\hat G_n \cdot \hat E_n \cdot \hat V_n - \alpha)
    \big)
\end{equation}
where $\hat G_n = G_n^{\lambda_g}, \hat E_n = -\log(\lambda_e E_n + \varepsilon_e)$, and $\hat V_n = -\log(\lambda_v V_n + \varepsilon_v)$, with the hyperparameters $\lambda_{\text{ds}}, \lambda_g, \lambda_e, \lambda_v, \varepsilon_e, \varepsilon_v$, and $\alpha$. Similar to SplitGuard, the abundance of hyperparameters in \cite{fuFocusingPinocchioNose2023} and the lack of a learnable structure to them make the method fragile, and its generalization to unseen scenarios becomes harder.

In the end, it has been shown that earlier methods can achieve almost-perfect accuracy in detecting training-hijacking attacks given the correct setting, but they give up on the clients' compute capabilities in favor of heuristic-driven computations. Therefore, the applicability of a less fragile and successful out-of-the-box method is required. Our main purpose in this work is to demonstrate that simpler methods can be utilized instead when clients are assumed to have modest computing power, which is reasonable as data sharing more than efficiency motivates collaborative ML setups in many scenarios. 

\section{SplitOut}

The core idea behind SplitOut is for the clients to train the entire network for a short duration using a small fraction (such as $1\%$) of their private data. The gradients collected will then be used as training data for an outlier detection method, which we expect will flag the gradients from an honest server as inliers and those from a training-hijacking server as outliers. The method as we will describe below can be implemented by using an out-of-the-box outlier detection method from libraries such as scikit-learn \cite{scikit-learn} and requires only the addition of small amounts of bookkeeping code on top an existing SplitNN training setup. By additionally integrating a window-based detection on top of an existing outlier detection algorithm, we further improve our true and false positive rates.

\subsection{Local Outlier Factor (LOF)}

An outlier detection (OD) method tries to identify data points that are ``different" from the ``normal," e.g. to identify potentially dubious financial transactions. Local outlier factor (LOF) \cite{breunig2000lof, chen2010comparison, janssens2009outlier} is an unsupervised OD method readily implemented in popular libraries such as scikit-learn \cite{scikit-learn}. It does not make any distribution assumptions and is density-based. Rather than outputting a binary decision, LOF assigns a \textit{local outlier factor score} to each point, where points in homogeneous clusters have low scores and those with smaller local densities have high scores. The LOF algorithm works as follows:

Using a distance measure $d(\cdot,\cdot)$ such as the Euclidean distance, calculate the \textit{k-distance} (the distance between a point $p$ and its $k^{th}$ closest neighbor $p_k$), denoted $d_k(p)$ of each point $p$. Then calculate its \textit{reachability distance} to each point $q$ as
\begin{equation}
\label{eqn:reach-dist}
    r(p, q) = \max(d(p,q), d_k(p)),
\end{equation}
and its \textit{local reachability density(LRD)} as the inverse of its average reachability distance to its $k$ nearest neighbors $\text{kNN}(p)$:
\begin{equation}
\label{eqn:lrd}
    \text{LRD}(p) = \left(\frac{1}{k} \sum_{q \in \text{kNN(p)}} r(p, q)\right)^{-1}.
\end{equation}
Finally, assign a LOF score to $p$ as the ratio between the average LRD of its $k$-neighbors and its own LRD:
\begin{equation}
\label{eqn:lof-score}
    \text{LOF}(p) = \frac{\sum_{q \in \text{kNN(p)}} \text{LRD}(q)}{k \cdot \text{LRD(p)}}.
\end{equation}

According to the equation \ref{eqn:lrd}, local reachability density describes the density of other points around the selected point. In other words, if the closest points cluster does not have a relatively small distance from the selected point, LRD produces small values. As stated in Equation \ref{eqn:lof-score}, the LRD values calculated for each point are compared with the average LRD of the k nearest neighbors of the selected point. For the point where the inlier detection is desired, the average LRD of neighbors is expected to be equal to or approximately close to the LRD of the selected point. Finally, we expect that if the point $p$ is an inlier, $\text{LOF}(p)$ will be nearly 1 and greater otherwise. Generally, we say $p$ is an outlier if $\text{LOF}(p) > 1$. 

Since LOF discovers outliers only based on their local densities and does not try to model the distribution, it does not require us to have even a rough model of the expected outlier behavior, which makes it a feasible choice: One does not need to know the attacker's behavior beforehand. Hence, it forms the basis of our detection methodology.

\subsection{Collecting Training Data for Outlier Detection}

The clients collect training data for LOF by training the entire network for a short time (e.g. one epoch) using a small fraction (such as $1\%$) of their data. We thus assume that the clients have access to the whole neural network architecture but not necessarily the same parameters with the server-side model. They can then train the neural network for one epoch with 1\% of their data \textit{before the actual training} to collect a set of \textit{honest} gradients.

\begin{algorithm}[t!]
    \caption{SplitOut: Client Training}
    \begin{algorithmic}[1]
        {\color{gray}
        \STATE $f_1, f_2$: client and server models \\ 
        \STATE $f_2^{\text{SIM}}$: simulated server model \\
        \STATE $S$: fct. to get subset (e.g. 1\%) of the dataset \\
        \STATE $k$: no. of neighbors for LOF \\
        \STATE $w$: LOF decision window size \\ 
        \STATE $train_{\text{LOF}}$: gradients to train LOF \label{lst:line:train-lof-grads}\\
        \STATE $pred_{\text{LOF}}$: gradients for LOF to classify \\ }
        \STATE initialize $train_{\text{LOF}}$ $pred_{\text{LOF}}$ as empty lists \\
        
        \STATE{\color{gray}// simulating local model to train LOF} \\
        
        \FOR{$(x_i,y_i)$ $\leftarrow$ $S(\text{trainset})$} 
            \STATE Train $f_1$ and $f_2^{\text{SIM}}$ on $(x_i, y_i)$ \label{lst:line:lof-training} \\ 
            \STATE Append $f_1$'s gradients to $train_{\text{LOF}}$
        \ENDFOR 

        \STATE {\color{gray}// actual SplitNN training with server} \\
        \FOR{$(x_i,y_i)$ $\leftarrow$ trainset} 
            \STATE Send $f_1(x_i)$ and (optional) $y_i$ to the server \\
            \STATE Obtain gradients $\nabla_1$ from the server \label{lst:line:get-grads} \\
            \STATE $pred_{\text{LOF}} \leftarrow \nabla_1$ \\
            \STATE Remove older values from $pred_{\text{LOF}}$ to keep its size fixed at $w$. \\
            \IF{$|pred_{\text{LOF}}| < w$} \label{lst:line:check-window}
              \STATE \textbf{Continue} to next iteration.
            \ENDIF
            \IF{LOF$(train_{\text{LOF}}, k)$ classifies majority of $pred_{\text{LOF}}$ as outliers} \label{lst:line:check-attack}
              \STATE \textbf{Return} ``attack".
            \ENDIF
        \ENDFOR

    \end{algorithmic}
    \label{alg:ad_cli}
\end{algorithm}

\begin{figure*}[]
    \centering
    \makebox[\textwidth][c]{\includegraphics[width=1\textwidth]{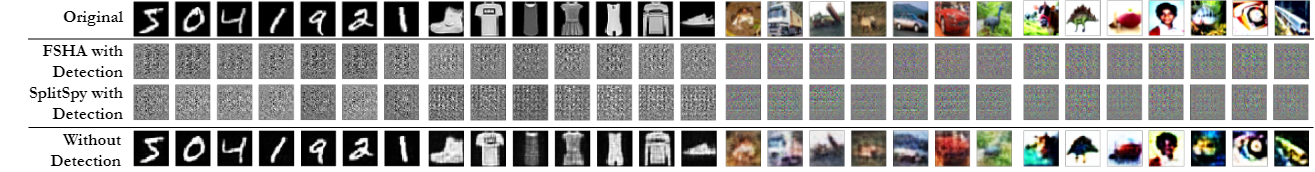}}%
    \caption{\textbf{Results obtained by attackers} for the MNIST, F-MNIST, CIFAR10, and CIFAR100 datasets with respect to the detection times (as shown in Table \ref{tab:detection_results_v2}). The first row displays the original images, and the last row displays the results obtained by a FSHA \cite{pasquini_unleashing_2021} attacker able to run for an arbitrary duration without being detected.}
    \label{fig:fsha_results}
\end{figure*}

\begin{table*}[]
     \centering
     \caption{Attack detection statistics using SplitOut, collected over 100 runs of the first epoch of training with FSHA \cite{pasquini_unleashing_2021}, the backdoor attack \cite{yuHowBackdoorSplit2023}, SplitSpy \cite{fuFocusingPinocchioNose2023}, FSHA with multitask learning (FSHA-MT, returning the average of honest and adversarial losses), and an honest server.}
    \label{tab:detection_results_v2}
        \begin{tabular}{@{}lccccccccc@{}}
            \toprule
                             & \multicolumn{2}{l}{ \textbf{ FSHA } }  & \multicolumn{2}{l}{ \textbf{ SplitSpy } } & \multicolumn{2}{l}{ \textbf{ Backdoor } } & \multicolumn{2}{l}{ \textbf{ FSHA-MT } } &              \\ \midrule
            & \textbf{TPR} & \textit{\textbf{t}} & \textbf{TPR}   & \textit{\textbf{t}}  & \textbf{TPR}   & \textit{\textbf{t}}  & \textbf{TPR}  & \textit{\textbf{t}}  & \textbf{FPR} \\ \midrule
            MNIST            & 1          & .011               & 1            & .011                & 1            & .011                & 1.00          & .012                & 0            \\
            F-MNIST          & 1          & .011               & 1            & .011                & 1            & .011                & 0.98          & .029                & 0            \\
            CIFAR10          & 1          & .013               & 1            & .013                & 1            & .013                & 0.98          & .026                & .02         \\
            CIFAR100         & 1          & .013               & 1            & .013                & 1            & .013                & 1.00          & .064                & .14         \\ \bottomrule
        \end{tabular}
\vspace{-10pt}       
\end{table*}

This seemingly contradicts the goal of collaborative ML, which was to outsource computation. However, as we will demonstrate in more detail, clients using a very small share of their local data for the data collection phase still perform accurate detection. Moreover, better data utilization, rather than purely outsourcing computation, might be a stronger reason for using SplitNN in certain scenarios (e.g. financially/technologically capable institutions such as universities or hospitals working on regulated private data, as exemplified in \cite{poirot2019split}). Since to collect training data for LOF we need only one epoch of training with 1\% of the whole data, this is still feasible for many scenarios where SplitNN is employed. Recall that in SplitNN, the clients would train several layers (but not the whole neural network) with 100\% of their data for multiple epochs, and increasing the number of client-side layers is recommended to make the private data more difficult to reconstruct from the intermediate outputs \cite{erdogan2021unsplit}.

\subsection{Detecting the Attack}

When the SplitNN training begins, the gradients the server sends are input to LOF, and are classified as \textit{honest} (inlier) or \textit{malicious} (outlier). Clients can make a decision after each gradient or combine the results from multiple gradients, by classifying the last few gradients and reaching the final decision by a majority vote between them. If a client concludes that the server is launching an attack, it can stop training and prevent further progress of the attack.

For the LOF algorithm, clients need to decide on a hyperparameter $k$, the number of neighbors to consider. We have followed the existing work on automatically choosing the value of $k$ \cite{xu2019automatic}, and observed that setting it equal to one less than the number of honest gradients collected (i.e. the highest feasible value) consistently achieves the best performance across all our datasets. 

Algorithm \ref{alg:ad_cli} explains the attack detection process in detail. Before training, clients allocate a certain share of their training data as a separate set $train_{\text{SIM}}$ and train the client-side layers together with a local copy of the server-side layers using that dataset (lines 10-13). The resulting gradients obtained from server-side layers then constitute the training data for outlier detection, denoted $train_{\text{LOF}}$ (line 12). During the actual training with SplitNN, clients receive the gradients from the server as part of the SplitNN protocol (line 17); each gradient received from the server is input to LOF (line 18), and classified as an outlier/inlier (lines 23-25). With the window size $w$, clients decide that the server is attacking if at least $w$ LOF decisions has been collected and the majority of them are outliers (i.e. malicious gradients). Simple bookkeeping is done to remove the earlier gradients so that the prediction set remains at size $w$ (line 19), and no decision is made for the first $w-1$ iterations (lines 20-22).  Training is stopped as soon as an attack is detected. 

It is possible to employ another outlier detection method in SplitOut besides LOF. We also experimented with other method such as one-class SVM \cite{scholkopf1999support} but found results employing LOF to be far superior, and hence present SplitOut and results as using LOF. Experimental results with one-class SVM are shown in the Appendix \ref{detection-oc-svm}.

\section{Evaluation \& Discussion}

To evaluate our method, we attempt to answer the following reserach questions (RQs):

\begin{description}
    \item[RQ1] How accurately can SplitOut detect training-hijacking attacks of different kinds? (Section \ref{sec:rq1})
    \item[RQ2] How accurately can SplitOut detect training-hijacking attacks performed by a multi-tasking attacker learning both the adversarial and honest objectives? (Section \ref{sec:rq2_3})
    \item[RQ3] How accurately can SplitOut detect training-hijacking attacks performed by the adaptive attacker of \cite{fuFocusingPinocchioNose2023}? (Section \ref{sec:rq2_3})
    \item[RQ4] What can we learn about the nature of training-hijacking attacks from the success of SplitOut? (Section \ref{sec_dim})
\end{description}


\begin{wraptable}{r}{0.5\textwidth}
\vspace{-25pt}
\centering
\caption{Detecting FSHA \cite{pasquini_unleashing_2021} using SplitOut for EMNIST-digits, averaged over 50 runs, with window size 10.}
\label{tab:emnist-window10}
\adjustbox{width=0.45\textwidth}{
\begin{tabular}{@{}c|ccc@{}}
\toprule
\textbf{ Data \% } & \textbf{TPR} & \textbf{t ($\pm$ SE)}     & \textbf{FPR  }   \\ \midrule
.10                     & 1          & $4.1 \cdot 10^{-5} \pm  0$       & 0         \\
.25                    & 1          & $4.1 \cdot  10^{-5} \pm 0$       & 0  \\
1                       & 1          & $4.1 \cdot 10^{-5} \pm  0$       & 0          \\
\bottomrule
\end{tabular}
}
\vspace{-20pt}
\end{wraptable}

\subsection{Setup}

For our experiments, we use the ResNet architecture \cite{he2015deep}, trained using Adam optimizer \cite{kingma_adam_2017} with learning rate of 0.001, on the MNIST \cite{lecun2010mnist}, Fashion-MNIST \cite{xiao2017/online}, CIFAR10/100 \cite{Krizhevsky09learningmultiple} datasets, and EMNIST \cite{cohen2017emnist} dataset. With a batch size of 64, one epoch is equal to 938 batches for MNIST and F-MNIST, 782 batches for CIFAR10/100, and 3750 batches for EMNIST-digits. We implemented SplitOut in Python (v 3.10) using the PyTorch (v 2.1.0) \cite{pytorch} and scikit-learn (v 1.2) \cite{scikit-learn} libraries. 

\begin{figure*}[t!]
    \centering
    \includegraphics[width=\textwidth]{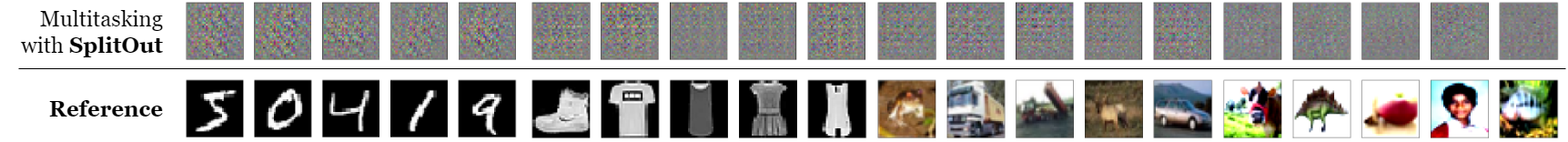}
    \caption{\textbf{Results the FSHA attacker obtains when it performs multitask learning until detection} for the MNIST, F-MNIST, CIFAR10, and CIFAR100 datasets, returning the gradients resulting from the average of honest training and adversarial objectives. The bottom row displays the original inputs.}
    \label{fig:mt_fsha_results}
\end{figure*}

We test SplitOut's performance when trained on with different shares of client training data (1\%, 10\%, 25\%, 50\%, 75\%, 100\%) and different decision-window sizes (1, 10). As stated earlier, we set the number of neighbors to the highest possible value (i.e. one less than the number of training points): 937 for MNIST/F-MNIST, 781 for CIFAR10/100, and 3749 for EMNIST-digits, when whole training data was used. If the batch size is expressed as $N_B$, the selected number of neighbors $k$ can be calculated as $(\lfloor N_B \rfloor -1)$.

Since MNIST and CIFAR datasets are close to each other in terms of total number of batches, the performance of SplitOut for EMNIST-digits dataset containing 3750 batches (with batch size 64) are analyzed. Apart from the selected data rates for other datasets, we evaluated the performance of SplitOut for EMNIST-digits dataset with relatively small data rates like 0.25\% (1\% of MNIST batches, 9 batches in total), and 0.1\% (only 3 batches).

\begin{wraptable}{r}{0.5\textwidth}
\vspace{-15pt}
\centering
\caption{FSHA detection statistics using SplitOut where client and server have different model architectures, averaged over 50 runs, with LOF data rate 75\% and window size 10.}
\label{tab:fsha-different-models}
\adjustbox{width=0.40\textwidth}{
\begin{tabular}{@{}lccc@{}}
\toprule
\textbf{Dataset} & \textbf{TPR} & \textbf{t ($\pm$ SE)}    & \textbf{FPR} \\ \midrule
MNIST            & 1          & $.0107 \pm .0000$  & 0          \\
F-MNIST          & 1          & $.0107 \pm .0001$  & 0          \\
CIFAR10          & 1          & $.0148 \pm .0003$ & .04         \\
CIFAR100         & 1          & $.0197 \pm .0038$    & .06 \\ \bottomrule        
\end{tabular}
}
\end{wraptable}

To test the consistency of the results in case the client cannot know the entire network, we also considered scenarios where the client and the honest server models have different architectures. While the client can deploy a model architecture of 4 ResNet blocks with 1,846,656 trainable parameters, SplitOut's performance was tested in the case where the actual split learning protocol has a model architecture of 6 ResNet blocks with 3,321,984 trainable parameters.

For each configuration, we measure the true and false positive rates by testing the method on 50 or 100 randomly-initialized runs each for the honest training scenario including training LOF from scratch, and each attack. A positive (attack detected) decision in a training-hijacking run is a true positive, and in an honest run is a false positive. To make sure we can detect the attack early enough, we start the algorithm after the required number of gradients for the decision window has been obtained.

\subsection{Detecting Training-Hijacking Attacks (RQ1)}
\label{sec:rq1}

The effectiveness of SplitOut against all split learning training-hijacking attacks proposed so far (FSHA \cite{pasquini_unleashing_2021}, backdoor attack \cite{yuHowBackdoorSplit2023}, SplitSpy \cite{fuFocusingPinocchioNose2023}) is displayed in Table \ref{tab:detection_results_v2}. We take the window size to be 10 and use as training data for LOF only $1\%$ of the clients' data, which corresponds to e.g. 9 and 7 training iterations in F-/MNIST and CIFAR10/100 datasets, respectively, with a batch size of 64. More comprehensive results covering a wider range of hyperparameters can be found in Appendix \ref{appx-comprehensive-results}.

\begin{figure}[t!]
    \centering
    \begin{subfigure}{0.40\linewidth}
        \centering
        \includegraphics[width=\textwidth]{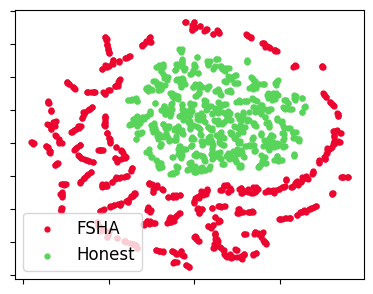}
        \caption{FSHA}
        \label{fig:tsne_fsha}
    \end{subfigure}  
    \begin{subfigure}{0.40\linewidth}
        \centering
        \includegraphics[width=\textwidth]{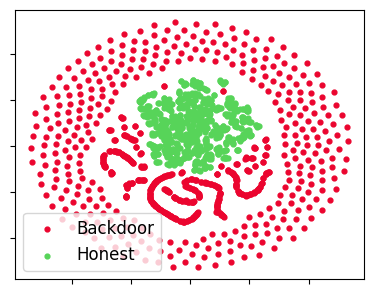}
        \caption{Backdoor Attack}
        \label{fig:tsne_backdoor}
    \end{subfigure}
    \caption{T-SNE \cite{van2008visualizing} dimension reduction comparing honest and malicious (FSHA \& the backdoor attack) gradients obtained from a randomly chosen run of the first epoch of training on CIFAR10.}
    \label{fig:tsne}
\end{figure}

In the more simpler MNIST and F-MNIST datasets, SplitOut achieves a true positive rate (TPR) of 1 and a false positive rate (FPR) of 0, meaning it can detect whether there is an attack or not with perfect accuracy. For the more complex CIFAR datasets, $2\%$ $\sim$ $14\%$ of honest runs are classified as malicious, but that also reduces to zero as more data is allocated for LOF training (see Appendix \ref{appx-comprehensive-results}), resulting in a higher compute load for the clients. In the dataset with a larger number of batches like EMNIST-digits, SplitOut can detect all attacks even with very small data rates such as 0.1\%, 0.25\%, and can also differentiate non-attack situations. Importantly, LOF's time to detect attacks does not vary significantly among the different training-hijacking attacks presented so far. 

In a scenario where the client and server have different model architectures and the client cannot know the actual architecture exactly, SplitOut can detect all attacks for all 4 datasets. However, due to reasons such as the client's model being smaller and the number of parameters being less, LOF should be trained with more data (for the rates we used in our experiments, more than 50\%) in order for the false positive rate rate to approach 0 in the CIFAR datasets. When evaluating the success of LOF in this scenario, it should be taken into account that the client has a smaller model in our setup.

\begin{table}[t!]
    \centering
  \caption{\textbf{SSIM \cite{wang-ssim} values} comparing reconstructed images by the attackers with the original private image averaged over 10 runs. The bottom row displays SSIM results from FSHA's reconstructed images without detection. A more successful reconstruction process is implied by a higher SSIM value.}
    \label{tab:ssim_results}
    \begin{adjustbox}{width=0.8\columnwidth,center}
    \begin{tabular}{@{}lcccc@{}}
    \toprule
    & \multicolumn{1}{l}{\textbf{MNIST}} & \textbf{F-MNIST} & \textbf{CIFAR} & \textbf{CIFAR100} \\ \midrule
    FSHA w/ SplitOut              & $.0003\pm .0007 $                            & $.0006\pm .0010$           & $.0006\pm .0008$         & $.0004\pm .0002 $           \\
    SplitSpy w/ SplitOut          & $.0001\pm .0004 $                            & $.0002\pm .0014$           & $.0005\pm .0011$         & $.0002\pm .0002 $           \\ \midrule
    Without detection             & $.8455\pm .0162$                             & $.6770\pm .0199$           & $.5189\pm .0317$         & $.2339\pm .0317$            \\ \bottomrule
    \end{tabular}
    \end{adjustbox}
    \end{table}

The structural similarity index measure (SSIM) \cite{wang-ssim} is a metric used to compare the quality of two different images. In our study, we used SSIM to estimate the image quality of the image reconstructed by the attacker, taking the original input image as a reference. For each dataset, SSIM values of 10 reconstructed images were calculated using the original image as reference, and their averages are shown in Table \ref{tab:ssim_results}. As previously shown in Figure \ref{fig:fsha_results} and indicated by the low SSIM values in Table \ref{tab:ssim_results}, SplitOut detects different attacks very early, preventing the malicious server from performing successful image reconstruction. 

\subsection{Detecting Adaptive Attacks (RQs 2,3)}
\label{sec:rq2_3}

To try to overcome a detection mechanism, a training-hijacking attacker can perform \textit{multitask learning} and combine the feature-space alignment and honest training losses, e.g. by taking their average, and returning the resulting gradients to the client model. To demonstrate this scenario, we consider a FSHA attacker  returning the average of the two losses. Table \ref{tab:detection_results_v2} displays our results against a multitasking FSHA (denoted FSHA-MT) attacker as well across our four benchmark datasets. We can detect all instances of the attack on MNIST and CIFAR100 datasets, and 98\% of the attack instances on F-MNIST and CIFAR10. Furthermore, Figure \ref{fig:mt_fsha_results} displays the reconstruction results the attacker achieves under this setting. Compared to the results in Figure \ref{fig:fsha_results}, the attack is expectedly less effective, since the client model is now updated to optimize the honest training loss as well as the adversarial discriminator loss. 

To show that SplitOut can be used cooperatively with the earlier detection work, we also consider the scenario demonstrated in \cite{fuFocusingPinocchioNose2023}, where the training-hijacking attacker has full knowledge of the detection mechanism in \cite{splitguard}. We use our approach in tandem with the method of \cite{splitguard}, and consider only the non-randomized batches as inputs to LOF. We use the name \textit{SplitSpy} for this approach following \cite{fuFocusingPinocchioNose2023} with the detection results displayed in Table \ref{tab:detection_results_v2}, where the LOF algorithm detects all instances of the attack similar to the earlier scenarios. This demonstrates that different detection methods can be used together and that even if the attacker circumvents the detection method of \cite{splitguard}, the attack can still be detected using SplitOut.

\subsection{Avoiding the Curse of Dimensionality (RQ4)} \label{sec_dim}

LOF is a distance-based method that makes frequent use of nearest-neighbor computations. Since the data we are working with is very high dimensional, the \textit{curse of dimensionality} could have an adverse impact and lead to poor performance. More formally, it has been shown in \cite{beyer1999nearest} that for all $L_p$ norms with $p \geq 1$, as the dimensionality increases, the proportion $(D_{\text{max}} - D_{\text{min}}) / D_{\text{min}}$ with $D_{\text{max}}$ the distance to the farthest point and $D_{\text{min}}$ to the nearest point vanishes, meaning that the concept of a neighborhood basically collapses. Although $L_p$ norms with small $p$ values have been shown to be preferable in higher dimensions \cite{aggarwalSurprisingBehaviorDistance2001}, they are still subject to the curse of dimensionality \cite{zimekSurveyUnsupervisedOutlier2012}.

\begin{figure*}[t!]
    \centering
    \includegraphics[width=0.9\textwidth]{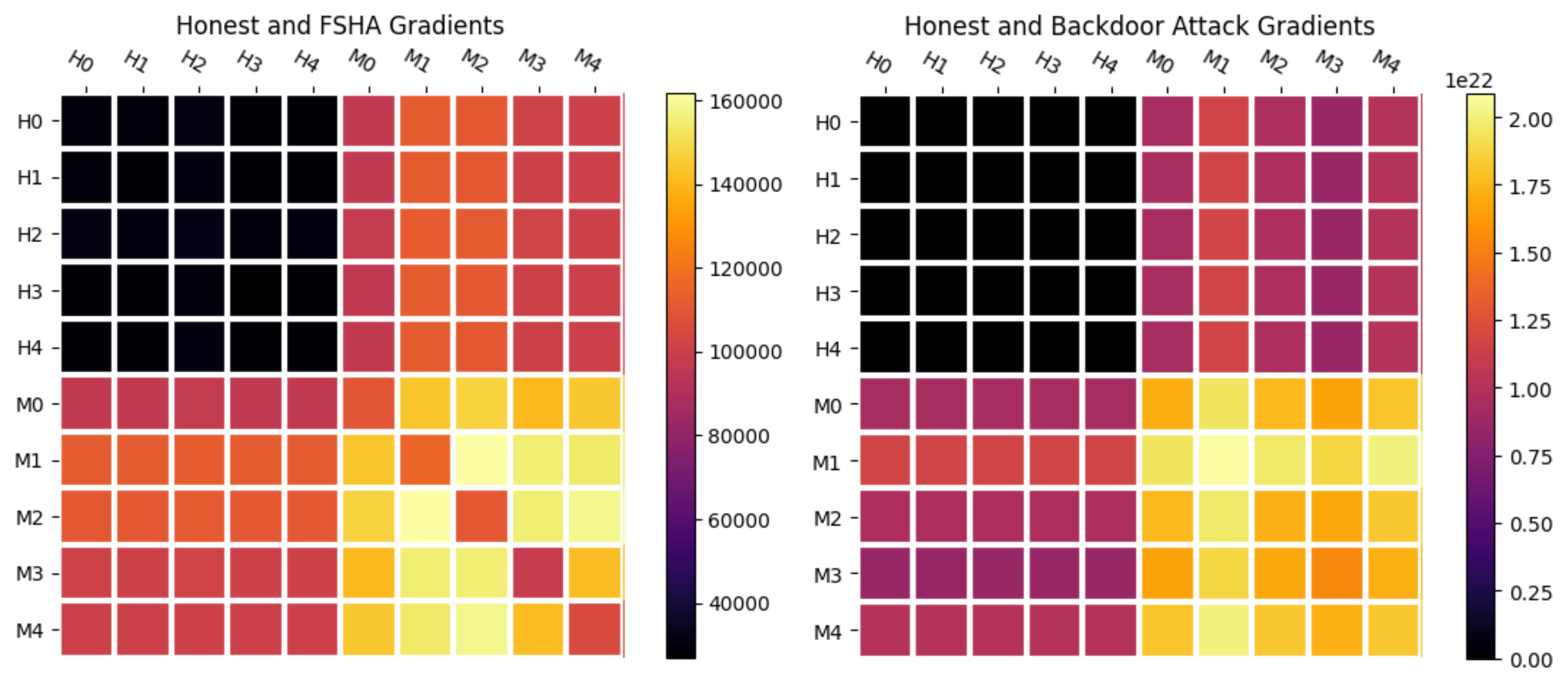}
    \caption{Pair-wise average $L_2$ distances between five randomly-chosen sets of gradients (corresponding to the first epoch) for honest training, FSHA \cite{pasquini_unleashing_2021} and the backdoor attack \cite{yuHowBackdoorSplit2023}, each using the CIFAR10 dataset. In each of the plots, $H$ indicates honest gradients and $M$ malicious. Darker colors imply smaller values.}
    \label{fig:distances}
\end{figure*}

In light of this the counter-intuitive accuracy of LoF requires an explanation. We now analyze the vast amount of data we collected throughout this work to explain the reasons behind the strong performance in a such high-dimensional setting. Figure \ref{fig:distances} displays the pair-wise average $L_2$ distances between five randomly-chosen sets of gradients from honest training, as well as FSHA \cite{pasquini_unleashing_2021} and the backdoor attack \cite{yuHowBackdoorSplit2023} all corresponding to the first epoch of training. The results are highly informative about the global and local structure of the problem:
\begin{itemize}
    \item \textit{Honest gradients are closer to honest gradients}, as the top left quadrant is darker than the top right (and its symmetric bottom left) quadrant.
    \item \textit{Honest gradients are also closer to each other among themselves than malicious gradients are} since the top left quadrant is darker than the bottom right quadrant. 
    \item \textit{Malicious gradients are closer to honest gradients than they are to other malicious gradients} since the bottom left (and top right) quadrants are darker than the bottom right quadrant. 
\end{itemize}
These three observations together imply that the neighborhoods of honest gradients are more likely to be densely populated with honest gradients, while malicious gradients have sparser neighborhoods that contain a number of honest gradients as well as malicious ones. 

The t-SNE \cite{van2008visualizing} plots (Figure \ref{fig:tsne}) displaying a dimension reduction of honest and malicious gradients over one epoch are consistent with the observations above since they also visualize honest gradients packed tightly, and malicious gradients covering a larger space essentially encompassing the honest gradients. This significant difference between honest and malicious neighborhoods is the main factor why a neighborhood-based method such as LOF succeeds in this task. 

So we conclude that while neighborhood-based methods generally struggle in high dimensions, the intrinsic differences between the honest training and feature-space alignment optimization processes result in them having significantly different forms to their gradient updates, which gives rise to neighborhoods structures easily distinguishable through the $L_2$ distance despite the high dimensionality.

\subsection{Computational Cost}



As stated earlier, the core assumption behind SplitOut is that the clients can reasonably perform forward/backward passes over the entire model for a very small number of batches (e.g. 9 in case of F-/MNIST and 7 in CIFAR10/100). It has also been demonstrated as part of the earlier data reconstruction attacks \cite{erdogan2021unsplit, pasquini_unleashing_2021} that higher number of client-side layers makes client outputs harder to invert, giving rise to the privacy/cost trade-off inherent in many privacy-preserving systems. Thus, unlike earlier work on detecting training-hijacking attacks \cite{splitguard, fuFocusingPinocchioNose2023} that relied on heuristical rather than computational assumptions as outlined in Section \ref{sec_detecting}, our work extends this privacy/cost trade-off by instead giving up on task-related assumptions. This leads in turn to a simpler-to-use detection method that can be applied out-of-the-box without any hyperparameter tuning.

\section{Related Work} \label{sec:related_work}
Split learning has been shown to be vulnerable to various kinds of attacks beyond the training-hijacking literature we have covered. Regarding data privacy, split learning is vulnerable to model inversion attacks \cite{erdogan2021unsplit, gao2023pcat, qiuEXACTExtensiveAttack2023, zhu2023passive} aiming to reconstruct private client data, and label leakage attacks \cite{erdogan2021unsplit, liu2022clustering, li2021label, kariyappa2023}, where the attacker (server) attempts to infer the private labels from the gradients sent by the client. Furthermore, confidentiality of the client models can be violated through model stealing attacks \cite{erdogan2021unsplit, gao2023pcat}, and the integrity of the system can be damaged through backdoor attacks \cite{bai2023villain}. 

Existence of such attacks has motivated the search for defense mechanisms. Differential privacy \cite{dwork2014algorithmic} has been applied to make private data reconstruction \cite{gawron2022feature, wu2023split} and label leakage \cite{yang2022differentially} harder. Further work has considered integrating an additional optimization step to minimize the information the intermediate outputs leak \cite{vepakomma_nopeek_2020}, utilizing the permutation-equivariance of certain deep neural network components by shuffling the inputs without a cost to the model performance \cite{xuShuffledTransformerPrivacyPreserving2023}, and using homomorphic encryption to compute part of the system on encrypted data to obtain formal guarantees on information leakage \cite{khan2023love}.

\section{Conclusion and Future Work}

We demonstrated that SplitOut, an out-of-the-box outlier detection approach using the local outlier factor algorithm, can detect whether a split learning server is launching a training-hijacking attack or not with almost-perfect accuracy. Compared to earlier such detection work, we make stronger assumptions on the clients' compute capabilities, but require fewer hyperparameters to be tuned, leading to a simpler to method that can be implemented readily using popular frameworks such as scikit-learn \cite{scikit-learn} on top of an existing split learning system. Due to the proactive design of our solution (performing outlier detection against honestly trained gradients), it can potentially detect some not-yet-known attacks.

\textbf{Limitations.} A main limitation of SplitOut is that the space of training-hijacking attacks so far has been limited to \textit{feature-space alignment} attacks. While such attacks have demonstrated strong performance for data reconstruction and implementing backdoors, future attacks employing a different training-hijacking approach could potentially behave in a different way and make our approach less effective. Another limitation of SplitOut is its applicability when the attacks start at a later epoch after some amount of honest training. While all the attacks we have considered start with the beginning of training and their generalizability against client models of various convergence levels is not trivial, it is nevertheless reasonable that \textit{honest} gradients from a later epoch are classified as outliers by LOF trained with honest gradients from the first epoch. Since the previous detection methods \cite{splitguard, fuFocusingPinocchioNose2023} also have this limitation, we leave this as future work to potentially improve the field.

\section*{Acknowledgements}
We acknowledge the Scientific and Technological Research Council of Turkey (TÜBİTAK) project 119E088.

\printbibliography

\clearpage
\appendix

\setcounter{table}{0}
\setcounter{page}{1}
\setcounter{figure}{0}



\section{Attack Detection with One-class SVM} \label{detection-oc-svm}


\begin{wraptable}{r}{0.5\textwidth}
\vspace{-25pt}
\centering
\caption{\textbf{Detecting FSHA using One-class SVM \cite{pasquini_unleashing_2021}.} True and false positive rates for our method averaged over 100 runs. One-class SVM data rate is the percentage of client data used in simulating the entire model to obtain training data for the one-class SVM algorithm. Window size is the number of most recent single-gradient decisions to perform a majority vote over to reach the final decision. The $t$ field denotes the average point of detection (as the share of the total number of batches), and the standard error of the mean is shown. Parameter \textit{nu} was chosen as 0.0001, \textit{kernel} was chosen as ‘rbf’, and \textit{gamma} was chosen as auto' in one-class SVM model training.}

\label{tab:ocsvm-all}
\begin{subtable}{0.5\columnwidth}
    \caption{Window size: 1}
    \label{tab:oc-svm-selected-params-window1}
    \centering
    \adjustbox{width=0.9\textwidth}{
    \begin{tabular}{@{}c|ccc@{}}
    \toprule
    \textbf{Data Rate (\%) } & \textbf{ TPR} & \textbf{t}        & \textbf{FPR } \\ \midrule
    1                       & 1.0          & $0.0013\pm0.0$    & 1.00         \\
    10                      & 1.0          & $0.0013\pm0.0$    & 1.00         \\
    25                      & 1.0          & $0.0013\pm0.0$    & 1.00         \\
    50                      & 1.0          & $0.0013\pm0.0$    & 1.00         \\
    75                      & 1.0          & $0.0018\pm0.0002$ & 1.00         \\
    100                     & 1.0          & $0.0023\pm0.0002$ & 0.98         \\ \bottomrule
    \end{tabular}
    }
\end{subtable}

\vspace*{0.25 cm}

\begin{subtable}{0.5\columnwidth}
    \caption{Window size: 10}
    \label{tab:oc-svm-selected-params-window10}
    \centering

    \adjustbox{width=0.9\textwidth}{
    \begin{tabular}{@{}c|ccc@{}}
    \toprule
    \textbf{Data Rate (\%) } & \textbf{ TPR} & \textbf{t}     & \textbf{FPR } \\ \midrule
    1                       & 1.0          & $0.0128\pm0.0$ & 1.00         \\
    10                      & 1.0          & $0.0128\pm0.0$ & 1.00         \\
    25                      & 1.0          & $0.0128\pm0.0$ & 1.00         \\
    50                      & 1.0          & $0.0128\pm0.0$ & 1.00         \\
    75                      & 1.0          & $0.0128\pm0.0$ & 0.02         \\
    100                     & 1.0          & $0.0129\pm0.0$ & 0.00         \\ \bottomrule
    \end{tabular}
    }
    
\end{subtable}
\end{wraptable}

We experimented with one-class SVM \cite{scholkopf1999support} as another outlier detection model. However, one-class SVM, which is based on applying a prediction method to new data with a decision boundary, is sensitive to outliers and it can be shown as one of its few major disadvantages compared to LOF. Additionally, the number of parameters that need to be determined to train a successful one-class SVM model is considerably more compared to LOF, while LOF only needs the number of neighbors for training. 

In our experiments, we performed a hyperparameter search for \textit{kernel}, \textit{gamma} and \textit{nu} values to train the one-class SVM model, which we expected to keep the FPR value low while keeping the TPR value high. Since one-class SVM tries to draw boundaries for the existing data, and the \textit{kernel} parameter determines the kernel type to be used in the algorithm. Another parameter, \textit{gamma} coefficient should be selected for \textit{'rbf'}, \textit{'poly'} and \textit{'sigmoid'} kernels. Parameter \textit{nu} defines the upper bound of the ratio of outlier or training errors and the lower bound of the ratio of support vectors. Valid range of \textit{nu} for one-class SVM training should be (0,1].

We measured the performance of one-class SVM for FSHA \cite{pasquini_unleashing_2021} on CIFAR10 dataset with values for the kernel parameter (linear, poly, rbf, sigmoid, precomputed), for the gamma parameter (scale, auto), and for the nu parameter (0.0001, 0.001, 0.01, 0.1, 0.5).

As a result of our experiments, we observed that the most successful results were obtained when \textit{nu} was 0.0001, \textit{kernel} was ‘rbf’ and \textit{gamma} parameter was 'auto'. As can be seen in Table \ref{tab:ocsvm-all}, we observed that one-class SVM can approach the results of LOF only when the data rate is high (75\%, 100\%), while LOF can give the same results with low data rates such as 1\% and 10\%.



\begin{wraptable}{r}{0.5\textwidth}
\vspace{-25pt}
\centering
\caption{\textbf{Detecting FSHA \cite{pasquini_unleashing_2021} using SplitOut for EMNIST-digits dataset.} True and false positive rates for our method averaged over 50 runs. The $t$ field denotes the average point of detection (as the share of the total number of batches), and the standard error of the mean is shown. }
\label{tab:emnist-all}

\begin{subtable}{0.5\columnwidth}
    \caption{Window size: 1}
    \label{tab:emnist-all-window1}
    \centering

    \adjustbox{width=0.9\textwidth}{
    \begin{tabular}{@{}c|ccc@{}}
    \toprule
    \textbf{ Data Rate (\%) } & \textbf{ TPR } & \textbf{ $t$ }      & \textbf{ FPR } \\ \midrule
    0.1                     & 1          & $4.1x10^{-6} \pm 0.0$       & 1          \\
    0.25                    & 1          & $4.1x10^{-6} \pm 0.0$       & .98          \\
    1                       & 1          & $4.1x10^{-6} \pm 0.0$        & .98          \\
    10                      & 1          & $4.1x10^{-6} \pm 0.0$        & .94          \\
    25                      & 1          & $5.0x10^{-6} \pm 5.4x10^{-7}$ & .40          \\
    50                      & 1          & $9.9x10^{-6} \pm 1.3x10^{-6}$ & .04          \\
    75                      & 1          & $1.4x10^{-5} \pm 1.6x10^{-6}$ & 0          \\
    100                     & 1          & $1.6x10^{-5} \pm 1.6x10^{-6}$ & 0          \\ \bottomrule
    \end{tabular}
    }
\end{subtable}

\vspace*{0.25 cm}

\begin{subtable}{0.5\columnwidth}
    \caption{Window size: 10}
    \label{tab:emnist-all-window10}
    \centering

    \adjustbox{width=0.9\textwidth}{
    \begin{tabular}{@{}c|ccc@{}}
    \toprule
    \textbf{ Data Rate (\%) } & \textbf{ TPR } & \textbf{ $t$ }      & \textbf{ FPR } \\ \midrule
    0.1                     & 1          & $4.1x10^{-5} \pm 0.0$        & 0          \\
    0.25                    & 1          & $4.1x10^{-5} \pm 0.0$        & 0          \\
    1                       & 1          & $4.1x10^{-5} \pm 0.0$       & 0          \\
    10                      & 1          & $4.1x10^{-5} \pm 0.0$        & 0          \\
    25                      & 1          & $4.2x10^{-5} \pm 1.2x10^{-7}$ & 0          \\
    50                      & 1          & $4.3x10^{-5} \pm 4.2x10^{-7}$ & 0          \\
    75                      & 1          & $4.3x10^{-5} \pm 7.4x10^{-7}$ & 0          \\
    100                     & 1          & $4.4x10^{-5} \pm 7.4x10^{-7}$ & 0          \\ \bottomrule
    \end{tabular}
    }
\end{subtable}
\vspace{-10pt}
\end{wraptable}

\section{Comprehensive Results For Attack Detection} \label{appx-comprehensive-results}

In this section, we provide our ablation study results.
Clients collect training data for LOF by training the entire model over one epoch, with some portion of their data. During this training process, the impact of using their data at different fractions on attack detection was examined. SplitOut's performance against FSHA \cite{pasquini_unleashing_2021}, backdoor \cite{yuHowBackdoorSplit2023} and SplitSpy \cite{fuFocusingPinocchioNose2023} is shown respectively in Tables  \ref{tab:lof_backdoor_full_table}, \ref{tab:lof_splitspy_full_table}, and \ref{tab:lof_fsha_full_results}  when trained with different shares of client training data (1\%, 10\%, 25\%, 50\%, 75\%, 100\%) and different decision-window sizes (1, 10). The LOF data rate is the percentage of client data used in simulating the entire model to obtain training data for the LOF algorithm. Window size is the number of most recent single-gradient decisions to perform a majority vote over to reach the final decision.

The performance of Split Out against FSHA was also experimented for EMNIST-digits dataset. The results for the EMNIST-digits dataset containing 3750 batches with batch size 64 as used in all setups are shown in Table \ref{tab:emnist-all}.




\section{Comprehensive Results For Attack Detection in Later Epochs} \label{appx-comprehensive-results-fsha-later epoch}

\begin{figure*}[t!]
    \centering
    \includegraphics[width=\textwidth]{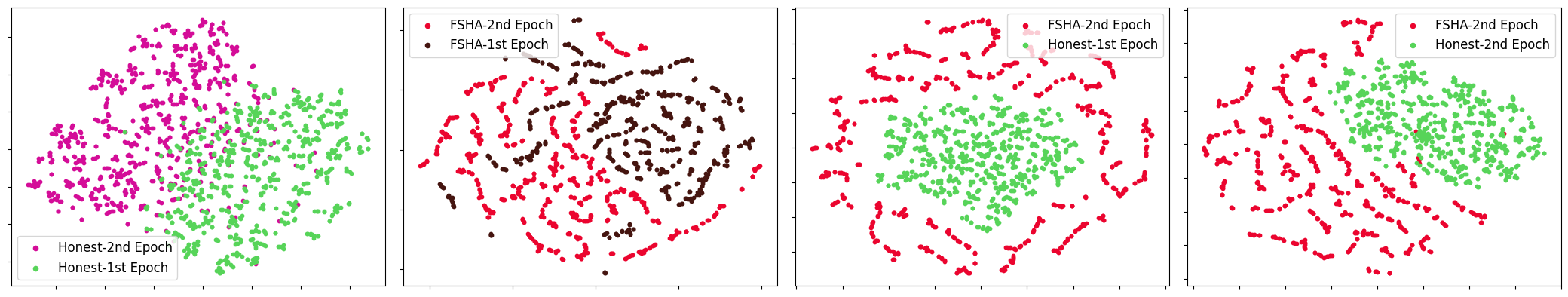},
    \caption{T-SNE \cite{van2008visualizing} dimension reduction comparing honest and malicious (FSHA) gradients obtained from a randomly chosen run of the first and second epoch of training on CIFAR10.}
    \label{fig:2ndepochtsne_horizontal}
\end{figure*}

To evaluate SplitOut's performance in later epochs, Table \ref{tab:fsha-later-epochs} includes detection statistics for when the attack starts in the second epoch. In this case, it is possible to completely detect the attack (with 0 FPR) with the SplitOut where the LOF model is trained with the first epoch gradients, although more data is needed than had the attack started in the first epoch, as can be seen in Table \ref{tab:fsha-later-epochs}. Besides, in Figure \ref{fig:2ndepochtsne_horizontal}, we added t-SNE plots to show the randomness and similarity properties of the honest and malicious gradients from the 2nd epoch and the 1st epoch gradients.

Although previously published defense methods \cite{splitguard, fuFocusingPinocchioNose2023} against training-hi-jacking have been created by considering the split learning protocol between the client and the honest server in cases where the attack is initiated simultaneously, the attacks do not strictly require starting against randomly-initialized clients. It is possible that an adaptive attack method can start in later epochs, but there is a trade-off for the attacker who trades attack time for more difficult detectability. Nevertheless, the attacks presented so far have started from the beginning of training, and their generalizability against client models of various convergence levels on different tasks is not trivial.

\begin{table}[ht!]
    \tiny
    \caption{\textbf{Detecting adaptive FSHA (in later epochs) using SplitOut \cite{pasquini_unleashing_2021}.} True and false positive rates for our method averaged over 50 runs. The $t$ field denotes the average point of detection (as the share of the total number of batches), and the standard error of the mean is shown.}
    \label{tab:fsha-later-epochs}
    \begin{subtable}{1\textwidth}
        \caption{Window size: 1}
        \label{tab:fsha-later-epochs-w1}
        \centering
        \makebox[\textwidth][c]{
        \begin{tabular}{@{}l|ccc|ccc|ccc@{}}
        \toprule
        \multicolumn{1}{c|}{\textbf{Method $\rightarrow$}}             & \multicolumn{3}{c|}{\textbf{ 1st Epoch Training }} & \multicolumn{3}{c|}{\textbf{ 2nd Epoch Training }} & \multicolumn{3}{c}{\textbf{ 1st+2nd Epoch Training }} \\ \midrule
        \multicolumn{1}{c|}{\textbf{Data \% $\downarrow$ }} & \textbf{ TPR }   & \textbf{FPR}   & \textbf{t}     & \textbf{ TPR }   & \textbf{FPR}   & \textbf{t}     & \textbf{ TPR }    & \textbf{FPR}    & \textbf{t}      \\ \midrule
        1                                                & 1            & 1            & $.0013\pm0$     & 1            & .64           & $.0076\pm.0014$  & 1             & 1            & $.0013\pm.0000$      \\
        10                                               & 1            & 1            & $.0013\pm0$     & .94           & .06           & $.0231\pm.0045$  & 1             & 1            & $.0013\pm.0001$   \\
        25                                               & 1            & 1            & $.0013\pm.0001$  & .92           & .02           & $.0309\pm.0056$  & 1             & .92            & $.0035\pm.0009$   \\
        50                                               & 1            & .92           & $.0035\pm.0009$  & .88           & .02           & $.0346\pm.0063$  & .96            & .10            & $.0212\pm.0047$   \\
        75                                               & .98           & .36           & $.0130\pm.0028$  & .82           & .00           & $.0886\pm.0278$  & .92            & .02            & $.0356\pm.0068$   \\
        100                                              & .96           & .10            & $.0212\pm.0047$  & .78           & .00           & $.0970\pm.0291$  & .80            & .00            & $.0932\pm.0285$   \\ \bottomrule
        \end{tabular}}
    \end{subtable}

    \vspace*{0.25 cm}
    \begin{subtable}{1\textwidth}
        \caption{Window size: 10}

        \centering
        \makebox[\textwidth][c]{
        \begin{tabular}{@{}l|ccc|ccc|ccc@{}}
        \toprule
        \multicolumn{1}{c|}{\textbf{Method $\rightarrow$}}             & \multicolumn{3}{c|}{\textbf{1st Epoch Training}} & \multicolumn{3}{c|}{\textbf{2nd Epoch Training}} & \multicolumn{3}{c}{\textbf{ 1st+2nd Epoch Training}} \\ \midrule
        \multicolumn{1}{c|}{\textbf{Data \% $\downarrow$}} & \textbf{ TPR }   & \textbf{FPR}   & \textbf{t}     & \textbf{ TPR }   & \textbf{FPR}   & \textbf{t}     & \textbf{ TPR }    & \textbf{FPR}    & \textbf{t}      \\ \midrule
        1                                                & 1            & 1           & $.0013\pm0$    & .96           & 0            & $.0452\pm.0093$  & 1             & 1            & $.00128\pm0$     \\
        10                                               & 1            & 1           & $.0013\pm0$    & .68           & 0            & $.0915\pm.0182$  & 1             & .84            & $.00128\pm0$     \\
        25                                               & 1            & .68           & $.0129\pm.0002$  & .44           & 0            & $.1179\pm.0257$  & 1             & .00            & $.0025\pm.0054$   \\
        50                                               & 1            & .00           & $.0250\pm.0054$  & .38           & 0            & $.1623\pm.0445$  & .78            & .00            & $.1042\pm.0178$   \\
        75                                               & .88           & .00           & $.0534\pm.0104$  & .28           & 0            & $.1568\pm.0524$  & .40             & .00            & $.1153\pm.0260$   \\
        100                                              & .78           & .00           & $.1042\pm.0178$  & .24           & 0            & $.1163\pm.0233$  & .28            & .00            & $.1595\pm.0520$    \\ \bottomrule
        \end{tabular}}
    \end{subtable}
\end{table}

\begin{table*}[]
    \tiny
    \caption{\textbf{Detecting FSHA using SplitOut \cite{pasquini_unleashing_2021}.} True and false positive rates for our method averaged over 100 runs. The $t$ field denotes the average point of detection (as the share of the total number of batches), and the standard error of the mean is shown.}
    \label{tab:lof_fsha_full_results}
    \begin{subtable}{1\textwidth}
        \caption{Window size: 1}
        \label{tab:lof_fsha_full_w1}
        \centering

        \makebox[\textwidth][c]{
        \begin{tabular}
        {l|ccc|ccc|ccc|ccc}
        \toprule
        \textbf{Dataset $\rightarrow$} & \multicolumn{3}{c}{\textbf{MNIST}} & \multicolumn{3}{|c|}{\textbf{F-MNIST}} & \multicolumn{3}{c|}{\textbf{CIFAR10}} & \multicolumn{3}{c}{\textbf{CIFAR100}} \\  \textbf{Data \% $\downarrow$}
                    & \, TPR & FPR   & $t$    & \, TPR & FPR & $t$ & \, TPR & FPR & $t$ & \, TPR & FPR & $t$  \\ \midrule
        1        & 1  & 0 & $.0013\pm0$  & 1  & 0 & $.0023\pm.0002$  & 1  & .98 & $.0019\pm.0002$  & 1  & .98 & $.0022\pm.0002$ \\
        10      & 1  & 0 & $.0013\pm0$   & 1  & 0 & $.0025\pm.0002$  & 1  & .96 & $.0019\pm.0002$ & 1  & .98 & $.0019\pm.0002$  \\ 
        25      & 1  & 0 & $.0013\pm0$   & 1  & 0 & $.0025\pm.0002$  & 1  & .52 & $.0044\pm.0003$ & 1  & .97 & $.0027\pm.0002$ \\ 
        50      & 1  & 0 & $.0013\pm0$   & 1  & 0 & $.0025\pm.0002$  & 1  & .03 & $.0078\pm.0002$ & 1  & .43 & $.0072\pm.0003$  \\
        75      & 1  & 0 & $.0013\pm0$   & 1  & 0 & $.0025\pm.0002$  & 1  & .00 & $.0093\pm.0003$ & 1  & .04 & $.0107\pm.0005$ \\
        100      & 1  & 0 & $.0013\pm0$   & 1  & 0 & $.0025\pm.0002$  & 1  & .00 & $.0186\pm.0011$ & 1  & .00 & $.0283\pm.0061$  \\
        \bottomrule
        \end{tabular}}
    \end{subtable}

    \vspace*{0.25 cm}
    
    \begin{subtable}{1\textwidth}
        \caption{Window size: 10}
        \label{tab:lof_fsha_full_w10}
        \centering
        \makebox[\textwidth][c]{
        \begin{tabular}{l|ccc|ccc|ccc|ccc}
        \toprule
        \textbf{Dataset $\rightarrow$} & \multicolumn{3}{c}{\textbf{MNIST}} & \multicolumn{3}{|c|}{\textbf{F-MNIST}} & \multicolumn{3}{c|}{\textbf{CIFAR10}} & \multicolumn{3}{c}{\textbf{CIFAR100}} \\  \textbf{Data \% $\downarrow$}
                    & \, TPR & FPR   & $t$    & \, TPR & FPR & $t$ & \, TPR & FPR & $t$ & \, TPR & FPR & $t$  \\ \midrule
        1        & 1  & 0 & $.0107\pm0$  & 1  & 0 & $.0109\pm.0001$  & 1  & .02 & $.0128\pm0$ & 1  & .14 & $.0129\pm0$  \\
        10      & 1  & 0 & $.0107\pm0$   & 1  & 0 & $.0109\pm.0001$  & 1  & .03 & $.0128\pm0$ & 1  & .41 & $.0129\pm0$  \\ 
        25      & 1  & 0 & $.0107\pm0$   & 1  & 0 & $.0109\pm.0001$  & 1  & .00 & $.0132\pm.0001$ & 1  & .14 & $.0129\pm0$ \\ 
        50      & 1  & 0 & $.0107\pm0$   & 1  & 0 & $.0109\pm.0001$  & 1  & .00 & $.0145\pm.0002$ & 1  & .00 & $.0145\pm.0002$  \\
        75      & 1  & 0 & $.0107\pm0$   & 1  & 0 & $.0109\pm.0001$  & 1  & .00 & $.0161\pm.0003$ & 1  & .00 & $.0190\pm.0011$  \\
        100      & 1  & 0 & $.0107\pm0$   & 1  & 0 & $.0109\pm.0001$  & .99  & .00 & $.0186\pm.0011$ & 1  & .00 & $.0808\pm.0081$  \\
      
        \bottomrule
        \end{tabular}}
    \end{subtable}
\end{table*}

\begin{table*}[]
    \tiny
    \caption{\textbf{Detecting the Backdoor Attack using SplitOut \cite{yuHowBackdoorSplit2023}.} True and false positive rates for our method averaged over 100 runs. The $t$ field denotes the average point of detection (as the share of the total number of batches), and the standard error of the mean is shown.}
    \label{tab:lof_backdoor_full_table}
    \begin{subtable}{1\textwidth}
        \caption{Window size: 1}
        \label{tab:lof_backdoor_full_w1}
        \centering

        \makebox[\textwidth][c]{
        \begin{tabular}
        {l|ccc|ccc|ccc|ccc}
        \toprule
        \textbf{Dataset $\rightarrow$} & \multicolumn{3}{c}{\textbf{MNIST}} & \multicolumn{3}{|c|}{\textbf{F-MNIST}} & \multicolumn{3}{c|}{\textbf{CIFAR10}} & \multicolumn{3}{c}{\textbf{CIFAR100}} \\  \textbf{Data \% $\downarrow$}
                    & \, TPR & FPR   & $t$    & \, TPR & FPR & $t$ & \, TPR & FPR & $t$ & \, TPR & FPR & $t$  \\ \midrule
        1        & 1  & 0 & $.0011\pm0$  & 1  & 0 & $.0011\pm0$  & 1  & .98 & $.0018\pm.0001$  & 1  & .98 & $.0014\pm.0001$  \\
        10      & 1  & 0 & $.0011\pm0$  & 1  & 0 & $.0011\pm0$  & 1  & .96 & $.0018\pm.0001$  & 1  & .98 & $.0014\pm.0001$  \\
        25      & 1  & 0 & $.0011\pm0$  & 1  & 0 & $.0011\pm0$  & 1  & .52 & $.0026\pm.0001$  & 1  & .97 & $.0015\pm.0001$  \\
        50      & 1  & 0 & $.0011\pm0$  & 1  & 0 & $.0011\pm0$  & 1  & .03 & $.0034\pm.0001$  & 1  & .43 & $.0051\pm.0003$  \\
        75      & 1  & 0 & $.0011\pm0$  & 1  & 0 & $.0011\pm0$  & 1  & .00 & $.0038\pm.0000$  & 1  & .04 & $.0083\pm.0004$  \\
        100      & 1  & 0 & $.0011\pm0$  & 1  & 0 & $.0011\pm0$  & 1  & .00 & $.0040\pm.0001$  & 1  & .00 & $.0176\pm.0043$  \\
        \bottomrule
        \end{tabular}}
    \end{subtable}

    \vspace*{0.25 cm}
    
    \begin{subtable}{1\textwidth}
        \caption{Window size: 10}
        \label{tab:lof_backdoor_full_w10}

        \centering
        \makebox[\textwidth][c]{
       \begin{tabular}
        {l|ccc|ccc|ccc|ccc}
        \toprule
        \textbf{Dataset $\rightarrow$} & \multicolumn{3}{c}{\textbf{MNIST}} & \multicolumn{3}{|c|}{\textbf{F-MNIST}} & \multicolumn{3}{c|}{\textbf{CIFAR10}} & \multicolumn{3}{c}{\textbf{CIFAR100}} \\  \textbf{Data \% $\downarrow$}
                    & \, TPR & FPR   & $t$    & \, TPR & FPR & $t$ & \, TPR & FPR & $t$ & \, TPR & FPR & $t$  \\ \midrule
        1        & 1  & 0 & $.0107\pm0$  & 1  & 0 & $.0107\pm0$  & 1  & .02 & $.0128\pm0$  & 1  & .14 & $.0128\pm0$  \\
        10      & 1  & 0 & $.0107\pm0$  & 1  & 0 & $.0107\pm0$  & 1  & .03 & $.0128\pm0$  & 1  & .41 & $.0128\pm0$  \\
        25      & 1  & 0 & $.0107\pm0$  & 1  & 0 & $.0107\pm0$  & 1  & .00 & $.0128\pm0$  & 1  & .14 & $.0128\pm0$  \\
        50      & 1  & 0 & $.0107\pm0$ & 1  & 0 & $.0107\pm0$  & 1  & .00 & $.0128\pm0$  & 1  & .00 & $.0134\pm.0001$  \\
        75      & 1  & 0 & $.0107\pm0$  & 1  & 0 & $.0107\pm0$  & 1  & .00 & $.0128\pm0$  & 1  & .00 & $.0173\pm.0019$ \\
        100      & 1  & 0 & $.0107\pm0$  & 1  & 0 & $.0107\pm0$  & 1  & .00 & $.0128\pm0$  & 1  & .00 & $.0308\pm.0043$  \\
        \bottomrule
        \end{tabular}}
    \end{subtable}
\end{table*}

\begin{table*}[ht!]
    \tiny
    \caption{\textbf{Detecting SplitSpy using SplitOut \cite{fuFocusingPinocchioNose2023}.} True and false positive rates for our method averaged over 100 runs. The $t$ field denotes the average point of detection (as the share of the total number of batches), and the standard error of the mean is shown.}
    \label{tab:lof_splitspy_full_table}
    \begin{subtable}{1\textwidth}
        \caption{Window size: 1}
        \label{tab:lof_splitspy_full_w1}
        \centering

        \makebox[\textwidth][c]{
        \begin{tabular}
        {l|ccc|ccc|ccc|ccc}
        \toprule
        \textbf{Dataset $\rightarrow$} & \multicolumn{3}{c}{\textbf{MNIST}} & \multicolumn{3}{|c|}{\textbf{F-MNIST}} & \multicolumn{3}{c|}{\textbf{CIFAR10}} & \multicolumn{3}{c}{\textbf{CIFAR100}} \\  \textbf{Data \% $\downarrow$}
                    & \, TPR & FPR   & $t$    & \, TPR & FPR & $t$ & \, TPR & FPR & $t$ & \, TPR & FPR & $t$  \\ \midrule
        1        & 1  & 0 & .0011$\pm$0  & 1  & 0 & .0012$\pm$.0001  & 1  & .98 & .0021$\pm$.0002  & 1  & .98 & .0023$\pm$.0002  \\
        10      & 1  & 0 & .0011$\pm$0  & 1  & 0 & .0012$\pm$.0001  & 1  & .96 & .0020$\pm$.0002  & 1  & .98 & .0019$\pm$.0002  \\
        25      & 1  & 0 & .0011$\pm$0  & 1  & 0 & .0012$\pm$.0001  & 1  & .52 & .0049$\pm$.0003  & 1  & .97 & .0026$\pm$.0002  \\
        50      & 1  & 0 & .0011$\pm$0  & 1  & 0 & .0012$\pm$.0001  & 1  & .03 & .0079$\pm$.0003  & 1  & .43 & .0068$\pm$.0003  \\
        75      & 1  & 0 & .0011$\pm$0  & 1  & 0 & .0012$\pm$.0001  & 1  & .00 & .0096$\pm$.0003  & 1  & .04 & .0108$\pm$.0004  \\
        100      & 1  & 0 & .0011$\pm$0  & 1  & 0 & .0012$\pm$.0001 & 1  & .00 & .0115$\pm$.0008  & 1  & .00 & .0195$\pm$.0033  \\
        \bottomrule
        \end{tabular}}
    \end{subtable}

    \vspace*{0.25 cm}
    
    \begin{subtable}{1\textwidth}
        \caption{Window size: 10}
        \label{tab:lof_splitspy_full_w10}
        \centering
        \makebox[\textwidth][c]{
        \begin{tabular}
        {l|ccc|ccc|ccc|ccc}
        \toprule
        \textbf{Dataset $\rightarrow$} & \multicolumn{3}{c}{\textbf{MNIST}} & \multicolumn{3}{|c|}{\textbf{F-MNIST}} & \multicolumn{3}{c|}{\textbf{CIFAR10}} & \multicolumn{3}{c}{\textbf{CIFAR100}} \\  \textbf{Data \% $\downarrow$}
                    & \, TPR & FPR   & $t$    & \, TPR & FPR & $t$ & \, TPR & FPR & $t$ & \, TPR & FPR & $t$  \\ \midrule
        1        & 1  & 0 & .0107$\pm$0  & 1  & 0 & .0107$\pm$0  & 1  & .02 & .0129$\pm$0  & 1  & .14 & .0129$\pm$0  \\
        10      & 1  & 0 & .0107$\pm$0  & 1  & 0 & .0107$\pm$0  & 1  & .03 & .0129$\pm$0  & 1  & .41 & .0129$\pm$0  \\
        25      & 1  & 0 & .0107$\pm$0  & 1  & 0 & .0107$\pm$0  & 1  & .00 & .0133$\pm$.0001  & 1  & .14 & .0129$\pm$0  \\
        50      & 1  & 0 & .0107$\pm$0  & 1  & 0 & .0107$\pm$0  & 1  & .00 & .0148$\pm$.0002  & 1  & .00 & .0144$\pm$.0002  \\
        75      & 1  & 0 & .0107$\pm$0  & 1  & 0 & .0107$\pm$0  & 1  & .00 & .0163$\pm$.0003  & 1  & .00 & .0206$\pm$.0029   \\
        100     & 1  & 0 & .0107$\pm$0  & 1  & 0 & .0107$\pm$0  & .98  & .00 & .0176$\pm$.0004  & 1  & .00 & .0422$\pm$.0072   \\
        \bottomrule
        \end{tabular}}
    \end{subtable}
\end{table*}

\begin{table*}[ht!]
    \tiny 
    \caption{\textbf{Detecting FSHA using SplitOut \cite{pasquini_unleashing_2021} where client and server have different models.} True and false positive rates for our method averaged over 50 runs. The $t$ field denotes the average point of detection (as the share of the total number of batches), and the standard error of the mean is shown.}
    \label{tab:lof_fsha_diffmodel_full_table}
    \begin{subtable}{1\textwidth}
        \caption{Window size: 1}
        \label{tab:lof_fsha_diffmodel_full_w1}
        \centering

        \makebox[\textwidth][c]{
        \begin{tabular}
        {l|ccc|ccc|ccc|ccc}
        \toprule
        \textbf{Dataset $\rightarrow$} & \multicolumn{3}{c}{\textbf{MNIST}} & \multicolumn{3}{|c|}{\textbf{F-MNIST}} & \multicolumn{3}{c|}{\textbf{CIFAR10}} & \multicolumn{3}{c}{\textbf{CIFAR100}} \\  \textbf{Data \% $\downarrow$}
                    & \, TPR & FPR   & $t$    & \, TPR & FPR & $t$ & \, TPR & FPR & $t$ & \, TPR & FPR & $t$  \\ \midrule
        1        & 1  & .84 & .0011$\pm$0  & 1  & .64 & .0011$\pm$0  & 1  & 1 & .0015$\pm$.0002  & 1  & 1 & .0014$\pm$.0001 \\
        10      & 1  & .84 & .0011$\pm$0   & 1  & .80 & .0011$\pm$0  & 1  & 1 & .0015$\pm$.0002 & 1  & 1 & .0014$\pm$.0001  \\ 
        25      & 1  & .84 & .0011$\pm$0   & 1  & .80 & .0011$\pm$0  & 1  & 1 & .0029$\pm$.0004 & 1  & 1 & .0016$\pm$.0002 \\ 
        50      & 1  & .84 & .0011$\pm$0   & 1  & .82 & .0011$\pm$0  & 1  & .98 & .0069$\pm$.0003 & 1  & 1 & .0050$\pm$.0004  \\
        75      & 1  & .84 & .0011$\pm$0   & 1  & .82 & .0011$\pm$0  & 1 & .88 & .0083$\pm$.0002 & 1  & .72 & .0088$\pm$.0007 \\
        100      & 1  & .84 & .0011$\pm$0   & 1  & .82 & .0011$\pm$0  & 1  & .76 & .0092$\pm$.0004 & 1  & .38 & .0121$\pm$.0011 \\
        \bottomrule
        \end{tabular}}
    \end{subtable}

    \vspace*{0.25 cm}
    
    \begin{subtable}{1\textwidth}
        \caption{Window size: 10}
        \label{tab:lof_fsha_diffmodel_full_w10}
        \centering
        \makebox[\textwidth][c]{
        \begin{tabular}{l|ccc|ccc|ccc|ccc}
        \toprule
        \textbf{Dataset $\rightarrow$} & \multicolumn{3}{c}{\textbf{MNIST}} & \multicolumn{3}{|c|}{\textbf{F-MNIST}} & \multicolumn{3}{c|}{\textbf{CIFAR10}} & \multicolumn{3}{c}{\textbf{CIFAR100}} \\  \textbf{Data \% $\downarrow$}
                    & \, TPR & FPR   & $t$    & \, TPR & FPR & $t$ & \, TPR & FPR & $t$ & \, TPR & FPR & $t$  \\ \midrule
        1        & 1  & 0 & .0107$\pm$0  & 1  & 0 & .0107$\pm$0  & 1  & 1 & .0128$\pm$0 & 1  & 1 & .0128$\pm$0  \\
        10      & 1  & 0 & .0107$\pm$0   & 1  & 0 & .0107$\pm$0  & 1  & .98 & .0128$\pm$0 & 1  & 1 & .0128$\pm$0  \\ 
        25      & 1  & 0 & .0107$\pm$0   & 1  & 0 & .0107$\pm$0  & 1  & .76 & .0130$\pm$.0001 & 1  & 1 & .0128$\pm$.0000 \\ 
        50      & 1  & 0 & .0107$\pm$0   & 1  & 0 & .0107$\pm$0  & 1  & .16 & .0139$\pm$.0002 & 1  & .50 & .0134$\pm$.0001  \\
        75      & 1  & 0 & .0107$\pm$0   & 1  & 0 & .0107$\pm$0  & 1  & .04 & .0148$\pm$.0003 & 1  & .06 & .0197$\pm$.0038  \\
        100      & 1  & 0 & .0107$\pm$0   & 1  & 0 & .0107$\pm$0  & 1  & .00 & .0159$\pm$.0005 & 1  & .02 & .0227$\pm$.0038  \\
      
        \bottomrule
        \end{tabular}}
    \end{subtable}
\end{table*}

\end{document}


\maketitle

\appendices

{\color{blue}
\section{Validating the Claims Underlying Label-Changing} \label{validation}

\setcounter{table}{0}
\setcounter{page}{1}
\setcounter{figure}{0}

XX demonstrate that fake gradients make a larger angle with regular gradients than the angle between two random subsets of regular gradients, and that fake gradients have a higher magnitude than regular gradients. For each dataset, Figures \ref{fig:angles} and \ref{fig:mags} display these values obtained during the first epoch of training with an honest server, averaged over 5 runs. 

Figure \ref{fig:angles} shows that $\theta(F,R)$ is consistently greater than $\theta(R_1, R_2)$. Note however that the difference is greater for MNIST (around $60^{\circ}$) than for Fashion-MNIST (around $30^{\circ}$) and the CIFAR datasets (around $10^{\circ}$). This means that as the model performs better (see Table \ref{tab:accuracies}), the difference between the angles becomes higher as well. 

Figure \ref{fig:mags} displays a similar relation between the $d(F,R)$ and $d(R_1, R_2)$ values obtained during the first epoch of training. For each of our datasets, $d(F,R)$ values are consistently higher than the $d(R_1, R_2)$ values, although the difference is smaller for CIFAR compared to MNIST for similar reasons with the angle values.

\begin{table}[h!]
    \centering
    \caption{\textbf{Test classification accuracies with SG-LC} of the ResNet model for the MNIST, F-MNIST, and CIFAR10/100 datasets for different $B_F$ values after three epochs of  training with SplitGuard, averaged over 10 runs with a $P_F$ of $0.1$.}
    \label{tab:accuracies}
    
    \begin{tabular}{ccccc}
    \toprule
        $B_F$ & \multicolumn{4}{c}{Classification Accuracy (\%)} \\ \midrule
                           & \textbf{MNIST} & \textbf{F-MNIST} & \textbf{CIFAR10} & \textbf{CIFAR100}\\ \midrule
        $0$ (Original)     & 98.68 & 89.60 & 64.24 & 36.5 \\ \midrule
        $8/64$             & 98.62 & 89.80 & 64.0 & 38.46 \\ 
        $16/64$            & 98.86 & 89.24 & 66.38 & 38.24 \\ 
        $32/64$            & 99.14 & 90.40 & 62.58 & 36.06 \\ 
        $64/64$            & 98.92 & 89.44 & 63.28 & 35.86 \\ 
    \bottomrule
    \end{tabular}
\end{table}

\section{Effect of Label-Changing on Model Performance} \label{model_performance}

Table \ref{tab:accuracies} displays the classification accuracy of the ResNet model on the test sets of our four benchmark datasets with different $B_F$ values after three epochs of training, averaged over 10 runs. The client model consists of a single convolutional layer, and the rest of the model is computed by the server. This is the worst-case scenario for this purpose, since the part of the model that is being updated with fake batches is as large as possible.

The results show that the model performs similarly when trained with and without SG-LC. There is not a noticeable and consistent decrease in performance for any of the datasets, even for high $B_F$ values such as 1. 

\begin{table}[h!]
    \caption{\textbf{Classification accuracies for the original model and the client's linear classifier} (averaged over 5 runs) for honest and random training scenarios. The linear classifier was introduced with the idea that perhaps its low performance could alert the client to an attack.}
    \centering
    \begin{tabular}{lcccc}
    \toprule
          & \multicolumn{4}{c}{Classification Accuracy (\%)} \\ \midrule
        \textbf{Training}  & \multicolumn{2}{c}{Honest} & \multicolumn{2}{c}{Random} \\ \midrule
        \textbf{Model}     & Original & Linear           & Original & Linear \\ \midrule
        MNIST              & 98.09    & 92.13   & 8.62     & 91.54 \\
        F-MNIST            & 86.13    & 81.53   & 9.40     & 83.18 \\
        CIFAR10            & 51.62    & 28.26   & 8.08     & 28.88 \\
        CIFAR100           & 22.92    & 9.26    & 0.94     & 8.86 \\
    \bottomrule
    \end{tabular}

    \label{tab:linear_classifier}
\end{table}

\section{Training-Hijacking and Learning a Classifier from the Clients' Outputs} \label{linear_classifier}

A reasonable argument against a training-hijacking detection protocol such as SplitGuard is that it might have been simpler (and practically impossible for the attacker to detect) if the clients, already knowing the label values, had attempted to train a classifier using their models' outputs and assessed its accuracy, the hypothesis being that training-hijacking would lead the clients towards outputting values from which it would be difficult to a train a classifier, whereas the values output by an honestly-trained client model would by necessity give rise to an accurate classifier.

We experimentally tested this argument for the worst-case scenario: the server learns random labels (effectively leading the clients towards outputting random values, rather than the FSHA server attempting to extract as much information as possible), and the client's classifier consists of a single linear layer. 

Table \ref{tab:linear_classifier} displays the original model's and the linear classifier's classification accuracies for the honest and random training scenarios for each of our four benchmark datasets averaged over five runs. 

In honest training, the linear classifier performs worse than the original, much more sophisticated, model. But most significantly, there is no noticeable difference between the linear classifier's performances in honest- and random-training scenarios; the difference is at most $1-2\%$, sometimes in the favor of the random-training linear classifer (see F-MNIST and CIFAR10 in Table \ref{tab:linear_classifier}). 

In short, attempting to train a linear classifier using the client's outputs does not provide a reliable detection mechanism against training-hijacking attacks.
}

\begin{table*}[ht!]
    \caption{\textbf{Detecting FSHA using SplitOut \cite{pasquini_unleashing_2021}.} True and false positive rates for our method averaged over 100 runs. LOF data rate is the percentage of client data used in simulating the entire model to obtain training data for the LOF algorithm. Window size is the number of most recent single-gradient decisions to perform a majority vote over to reach the final decision. The $t$ field denotes the average point of detection (as the share of total number of batches).}
    \label{tab:ad_full_results}
    \begin{subtable}{1\textwidth}
        \caption{Window size: 1}
        \label{tab:lof_fsha_full_table}
        \centering
        
        \begin{tabular}
        {l|ccc|ccc|ccc|ccc}
        \toprule
        \textbf{Dataset $\rightarrow$} & \multicolumn{3}{c}{\textbf{MNIST}} & \multicolumn{3}{|c|}{\textbf{F-MNIST}} & \multicolumn{3}{c|}{\textbf{CIFAR10}} & \multicolumn{3}{c}{\textbf{CIFAR100}} \\  \textbf{LOF Data Rate (\%) $\downarrow$}
                    & TPR & FPR   & $t$    & TPR & FPR & $t$ & TPR & FPR & $t$ & TPR & FPR & $t$  \\ \midrule
        1        & 1.0  & 0.03 & 0.0013  & 1.0  & 0.02 & 0.0023  & 1.0  & 0.58 & 0.0043  & 1.0  & 0.70 & 0.0590  \\
        10      & 1.0  & 0.03 & 0.0013   & 1.0  & 0.01 & 0.0023  & 1.0  & 0.07 & 0.0076 & 1.0  & 0.23 & 0.0111  \\ 
        25      & 1.0  & 0.03 & 0.0013   & 1.0  & 0.01 & 0.0023  & 1.0  & 0.01 & 0.0089 & 1.0  & 0.10 & 0.0129 \\ 
        50      & 1.0  & 0.03 & 0.0013   & 1.0  & 0.01 & 0.0023  & 1.0  & 0.01 & 0.0097 & 1.0  & 0.02 & 0.0195  \\
        75      & 1.0  & 0.03 & 0.0013   & 1.0  & 0.01 & 0.0024  & 1.0  & 0.01 & 0.0108 & 1.0  & 0.03 & 0.0265  \\
        100      & 1.0  & 0.03 & 0.0013   & 1.0  & 0.00 & 0.0024  & 1.0  & 0.01 & 0.0117 & 1.0  & 0.02 & 0.0354  \\
        \bottomrule
        \end{tabular}
    \end{subtable}

    \vspace*{0.25 cm}
    
    \begin{subtable}{1\textwidth}
        \caption{Window size: 10}

        \centering
        \begin{tabular}{l|ccc|ccc|ccc|ccc}
        \toprule
        \textbf{Dataset $\rightarrow$} & \multicolumn{3}{c}{\textbf{MNIST}} & \multicolumn{3}{|c|}{\textbf{F-MNIST}} & \multicolumn{3}{c|}{\textbf{CIFAR10}} & \multicolumn{3}{c}{\textbf{CIFAR100}} \\  \textbf{LOF Data Rate (\%) $\downarrow$}
                    & TPR & FPR   & $t$    & TPR & FPR & $t$ & TPR & FPR & $t$ & TPR & FPR & $t$  \\ \midrule
        1        & 1.0  & 0.0 & 0.0107  & 1.0  & 0.0 & 0.0109  & 1.0  & 0.02 & 0.0132  & 1.0  & 0.02 & 0.0142  \\
        10      & 1.0  & 0.0 & 0.0107   & 1.0  & 0.0 & 0.0109  & 1.0  & 0.00 & 0.0145 & 1.0  & 0.00 & 0.0219  \\ 
        25      & 1.0  & 0.0 & 0.0107   & 1.0  & 0.0 & 0.0109  & 1.0  & 0.00 & 0.0156 & 1.0  & 0.00 & 0.0289 \\ 
        50      & 1.0  & 0.0 & 0.0107   & 1.0  & 0.0 & 0.0109  & 1.0  & 0.00 & 0.0170 & 1.0  & 0.00 & 0.0357  \\
        75      & 1.0  & 0.0 & 0.0107   & 1.0  & 0.0 & 0.0109  & 0.99  & 0.00 & 0.0170 & 1.0  & 0.00 & 0.0434  \\
        100      & 1.0  & 0.0 & 0.0107   & 1.0  & 0.0 & 0.0109  & 0.99  & 0.00 & 0.0183 & 1.0  & 0.00 & 0.0557  \\
      
        \bottomrule
        \end{tabular}
    \end{subtable}
\end{table*}

\begin{table*}[ht!]
    \caption{\textbf{Detecting the Backdoor Attack using SplitOut \cite{yuHowBackdoorSplit2023}.} True and false positive rates for our method averaged over 100 runs. LOF data rate is the percentage of client data used in simulating the entire model to obtain training data for the LOF algorithm. Window size is the number of most recent single-gradient decisions to perform a majority vote over to reach the final decision. The $t$ field denotes the average point of detection (as the share of total number of batches).}
    \label{tab:ad_full_results}
    \begin{subtable}{1\textwidth}
        \caption{Window size: 1}
        \label{tab:lof_backdoor_full_table}
        \centering
        
        \begin{tabular}
        {l|ccc|ccc|ccc|ccc}
        \toprule
        \textbf{Dataset $\rightarrow$} & \multicolumn{3}{c}{\textbf{MNIST}} & \multicolumn{3}{|c|}{\textbf{F-MNIST}} & \multicolumn{3}{c|}{\textbf{CIFAR10}} & \multicolumn{3}{c}{\textbf{CIFAR100}} \\  \textbf{LOF Data Rate (\%) $\downarrow$}
                    & TPR & FPR   & $t$    & TPR & FPR & $t$ & TPR & FPR & $t$ & TPR & FPR & $t$  \\ \midrule
        1        & 1.0  & 0.03 & 0.0011  & 1.0  & 0.02 & 0.0011  & 1.0  & 0.58 & 0.0026  & 1.0  & 0.70 & 0.0028  \\
        10      & 1.0  & 0.03 & 0.0011  & 1.0  & 0.01 & 0.0011  & 1.0  & 0.07 & 0.0036  & 1.0  & 0.23 & 0.0039  \\
        25      & 1.0  & 0.03 & 0.0011  & 1.0  & 0.01 & 0.0011  & 1.0  & 0.01 & 0.0038  & 1.0  & 0.10 & 0.0041  \\
        50      & 1.0  & 0.03 & 0.0011  & 1.0  & 0.01 & 0.0011  & 1.0  & 0.01 & 0.0039  & 1.0  & 0.02 & 0.0044  \\
        75      & 1.0  & 0.03 & 0.0011  & 1.0  & 0.01 & 0.0011  & 1.0  & 0.01 & 0.0039  & 1.0  & 0.03 & 0.0045  \\
        100      & 1.0  & 0.03 & 0.0011  & 1.0  & 0.00 & 0.0011  & 1.0  & 0.01 & 0.0040  & 1.0  & 0.02 & 0.0046  \\
        \bottomrule
        \end{tabular}
    \end{subtable}

    \vspace*{0.25 cm}
    
    \begin{subtable}{1\textwidth}
        \caption{Window size: 10}

        \centering
       \begin{tabular}
        {l|ccc|ccc|ccc|ccc}
        \toprule
        \textbf{Dataset $\rightarrow$} & \multicolumn{3}{c}{\textbf{MNIST}} & \multicolumn{3}{|c|}{\textbf{F-MNIST}} & \multicolumn{3}{c|}{\textbf{CIFAR10}} & \multicolumn{3}{c}{\textbf{CIFAR100}} \\  \textbf{LOF Data Rate (\%) $\downarrow$}
                    & TPR & FPR   & $t$    & TPR & FPR & $t$ & TPR & FPR & $t$ & TPR & FPR & $t$  \\ \midrule
        1        & 1.0  & 0.0 & 0.0106  & 1.0  & 0.0 & 0.0107  & 1.0  & 0.02 & 0.0128  & 1.0  & 0.02 & 0.0128  \\
        10      & 1.0  & 0.0 & 0.0106  & 1.0  & 0.0 & 0.0107  & 1.0  & 0.00 & 0.0128  & 1.0  & 0.00 & 0.0128  \\
        25      & 1.0  & 0.0 & 0.0106  & 1.0  & 0.0 & 0.0107  & 1.0  & 0.00 & 0.0128  & 1.0  & 0.00 & 0.0128  \\
        50      & 1.0  & 0.0 & 0.0106  & 1.0  & 0.0 & 0.0107  & 1.0  & 0.00 & 0.0128  & 1.0  & 0.00 & 0.0128  \\
        75      & 1.0  & 0.0 & 0.0107  & 1.0  & 0.0 & 0.0107  & 1.0  & 0.00 & 0.0128  & 1.0  & 0.00 & 0.0128  \\
        100      & 1.0  & 0.0 & 0.0107  & 1.0  & 0.0 & 0.0107  & 1.0  & 0.00 & 0.0128  & 1.0  & 0.00 & 0.0128  \\
        \bottomrule
        \end{tabular}
    \end{subtable}
\end{table*}

\begin{table*}[ht!]
    \caption{\textbf{Detecting SplitSpy using SplitOut \cite{fuFocusingPinocchioNose2023}.} True and false positive rates for our method averaged over 100 runs. LOF data rate is the percentage of client data used in simulating the entire model to obtain training data for the LOF algorithm. Window size is the number of most recent single-gradient decisions to perform a majority vote over to reach the final decision. The $t$ field denotes the average point of detection (as the share of total number of batches).}
    \label{tab:ad_full_results}
    \begin{subtable}{1\textwidth}
        \caption{Window size: 1}
        \label{tab:lof_splitspy_full_table}
        \centering
        
        \begin{tabular}
        {l|ccc|ccc|ccc|ccc}
        \toprule
        \textbf{Dataset $\rightarrow$} & \multicolumn{3}{c}{\textbf{MNIST}} & \multicolumn{3}{|c|}{\textbf{F-MNIST}} & \multicolumn{3}{c|}{\textbf{CIFAR10}} & \multicolumn{3}{c}{\textbf{CIFAR100}} \\  \textbf{LOF Data Rate (\%) $\downarrow$}
                    & TPR & FPR   & $t$    & TPR & FPR & $t$ & TPR & FPR & $t$ & TPR & FPR & $t$  \\ \midrule
        1        & 1.0  & 0.03 & 0.001  & 1.0  & 0.02 & 0.0011  & 1.0  & 0.58 & 0.0052  & 1.0  & 0.70 & 0.0056  \\
        10      & 1.0  & 0.03 & 0.001  & 1.0  & 0.01 & 0.0011  & 1.0  & 0.07 & 0.0082  & 1.0  & 0.23 & 0.0101  \\
        25      & 1.0  & 0.03 & 0.001  & 1.0  & 0.01 & 0.0011  & 1.0  & 0.01 & 0.0100  & 1.0  & 0.10 & 0.1230  \\
        50      & 1.0  & 0.03 & 0.001  & 1.0  & 0.01 & 0.0011  & 1.0  & 0.01 & 0.0106  & 1.0  & 0.02 & 0.0171  \\
        75      & 1.0  & 0.03 & 0.001  & 1.0  & 0.01 & 0.0012  & 1.0  & 0.01 & 0.0102  & 1.0  & 0.03 & 0.0208  \\
        100      & 1.0  & 0.03 & 0.001  & 1.0  & 0.00 & 0.0012  & 1.0  & 0.01 & 0.0114  & 1.0  & 0.02 & 0.0227  \\
        \bottomrule
        \end{tabular}
    \end{subtable}

    \vspace*{0.25 cm}
    
    \begin{subtable}{1\textwidth}
        \caption{Window size: 10}

        \centering
        \begin{tabular}
        {l|ccc|ccc|ccc|ccc}
        \toprule
        \textbf{Dataset $\rightarrow$} & \multicolumn{3}{c}{\textbf{MNIST}} & \multicolumn{3}{|c|}{\textbf{F-MNIST}} & \multicolumn{3}{c|}{\textbf{CIFAR10}} & \multicolumn{3}{c}{\textbf{CIFAR100}} \\  \textbf{LOF Data Rate (\%) $\downarrow$}
                    & TPR & FPR   & $t$    & TPR & FPR & $t$ & TPR & FPR & $t$ & TPR & FPR & $t$  \\ \midrule
        1        & 1.0  & 0.0 & 0.0106  & 1.0  & 0.0 & 0.0107  & 1.00  & 0.02 & 0.0137  & 1.0  & 0.02 & 0.0173  \\
        10      & 1.0  & 0.0 & 0.0106  & 1.0  & 0.0 & 0.0107  & 1.00  & 0.00 & 0.0100  & 1.0  & 0.00 & 0.0199  \\
        25      & 1.0  & 0.0 & 0.0106  & 1.0  & 0.0 & 0.0107  & 0.99  & 0.00 & 0.0158  & 1.0  & 0.00 & 0.0250  \\
        50      & 1.0  & 0.0 & 0.0106  & 1.0  & 0.0 & 0.0107  & 0.99  & 0.00 & 0.0164  & 1.0  & 0.00 & 0.0347  \\
        75      & 1.0  & 0.0 & 0.0106  & 1.0  & 0.0 & 0.0107  & 1.00  & 0.00 & 0.0169  & 1.0  & 0.00 & 0.0423  \\
        100      & 1.0  & 0.0 & 0.0106  & 1.0  & 0.0 & 0.0107  & 0.99  & 0.00 & 0.0174  & 1.0  & 0.00 & 0.0456  \\
        \bottomrule
        \end{tabular}
    \end{subtable}
\end{table*}